\documentclass{article}
\usepackage{ijcai25}

\usepackage{times}
\usepackage{soul}
\usepackage{url}
\usepackage[hidelinks]{hyperref}
\usepackage[utf8]{inputenc}
\usepackage[small]{caption}
\usepackage{graphicx}
\usepackage{amsmath}
\usepackage{amsthm}
\usepackage{booktabs}
\usepackage{algorithm}
\usepackage[switch]{lineno}
\usepackage{algpseudocode}
\usepackage{amsfonts}
\usepackage{multirow}
\usepackage{colortbl}
\usepackage{booktabs}
\usepackage{svg}
\usepackage{array}

\title{Enhancing Table Recognition with Vision LLMs:\\ A Benchmark and Neighbor-Guided Toolchain Reasoner}

\author{
Yitong Zhou$^{1}$ \and
Mingyue Cheng$^{1}$\thanks{Corresponding author} \and
Qingyang Mao$^{1}$ \and
Jiahao Wang$^{1}$ \and
Feiyang Xu$^{2}$ \and
Xin Li$^{1,2}$ \\
\affiliations
$^1$State Key Laboratory of Cognitive Intelligence, University of Science and Technology of China\\
$^2$Artificial Intelligence Research Institute, iFLYTEK Co., Ltd\\
\{yitong.zhou, maoqy0503, jiahao.wang\}@mail.ustc.edu.cn,
\{mycheng, leexin\}@ustc.edu.cn,
fyxu2@iflytek.com
}

\begin{document}

\maketitle

\begin{abstract}

Pre-trained foundation models have recently made significant progress in table-related tasks such as table understanding and reasoning. However, recognizing the structure and content of unstructured tables using Vision Large Language Models (VLLMs) remains under-explored. To bridge this gap, we propose a benchmark based on a hierarchical design philosophy to evaluate the recognition capabilities of VLLMs in training-free scenarios. Through in-depth evaluations, we find that low-quality image input is a significant bottleneck in the recognition process. Drawing inspiration from this, we propose the \textbf{N}eighbor-\textbf{G}uided \textbf{T}oolchain \textbf{R}easoner (NGTR) framework, which is characterized by integrating diverse lightweight tools for visual operations aimed at mitigating issues with low-quality images. Specifically, we transfer a tool selection experience from a similar neighbor to the input and design a reflection module to supervise the tool invocation process. Extensive experiments on public datasets demonstrate that our approach significantly enhances the recognition capabilities of the vanilla VLLMs. We believe that the benchmark and framework could provide an alternative solution to table recognition\footnote{Our code is available at \url{https://github.com/lqzxt/NGTR}}.
\end{abstract}

\section{Introduction}

Tables are ubiquitous for organizing and communicating structured data across diverse domains, ranging from scientific literature and business reports to web pages and financial documents~\cite{ye2024closer,zheng2021global}. They store a wealth of information essential for applications such as knowledge discovery, decision support, and data-driven analytics~\cite{shwartz2022tabular,wang2024tabletime}. In the context of intelligent table applications, one fundamental yet challenging task is table recognition~\cite{zanibbi2004survey}: converting image-based table representations into structured data formats. Over the years, substantial efforts~\cite{salaheldin2024deep} have been made to address this problem, introducing various approaches to address challenges, such as image segmentation techniques and cell object detection methods.

\begin{figure}[t]
    \vspace{0.04in}
    \centering
    \includegraphics[width=0.48\textwidth]{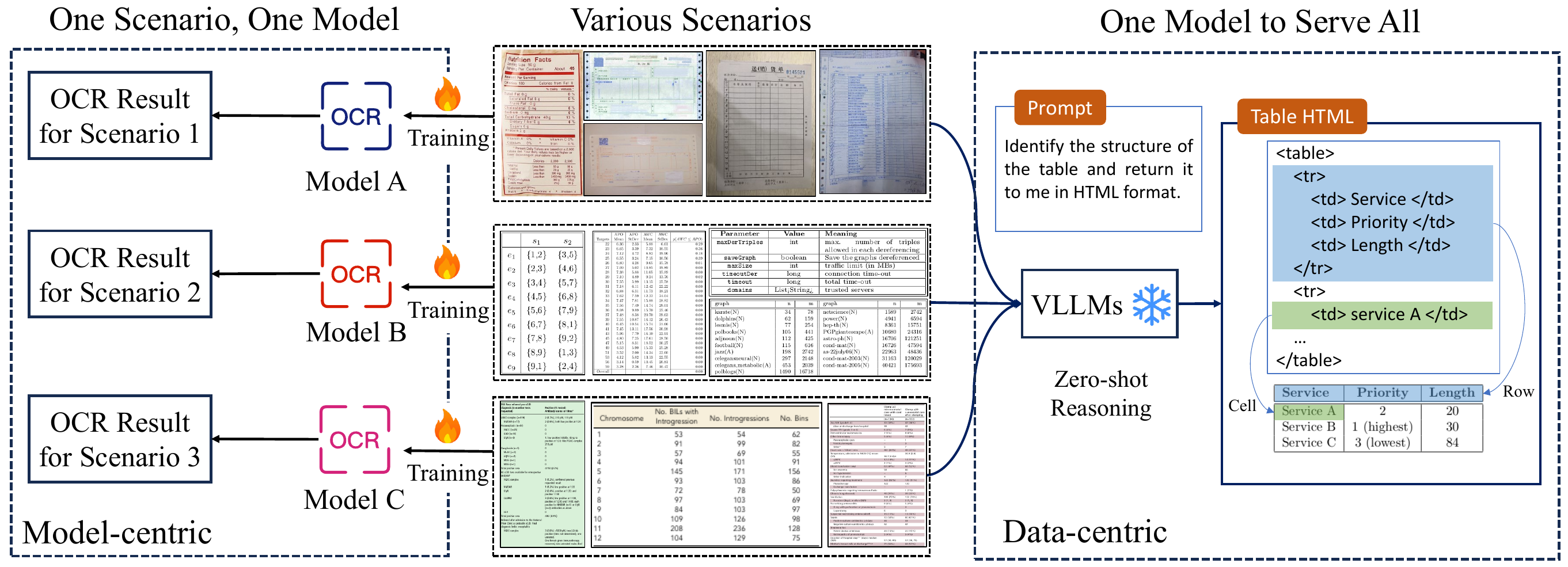}
    \vspace{-0.12in}
    \caption{Comparison of modeling paradigms: domain-specific lightweight models vs. universal pre-trained VLLMs.}
    \vspace{-0.15in}
    \label{motivation}
\end{figure}

Recently, the advent of Large Language Models (LLMs)~\cite{chang2024survey} and Vision Large Language Models (VLLMs)~\cite{yin2024survey} has revolutionized natural language processing and computer vision. For LLMs, their powerful understanding and reasoning capabilities have facilitated numerous tabular data mining tasks, such as table-to-text generation~\cite{guo2024adapting}, table question answering~\cite{wang2024chain,mao2024potable}, and table semantic understanding~\cite{deng2022turl,cheng2025survey2}.
Meanwhile, several VLLM-based methods have emerged to bypass traditional OCR pipelines for visual table analysis and understanding~\cite{hu2024mplug}. 

Despite these advancements, our investigation reveals a noticeable gap: the application of VLLMs to table recognition remains underexplored. This task serves as a foundational building block for table-related applications. Some existing work \cite{luo2024molar,zhang2024multi} has focused on pre-training or fine-tuning VLLMs to accomplish this task. However, fine-tuning VLLMs for specific tasks is often computationally expensive and risks catastrophic forgetting of general capabilities. To address this, we explore a generative approach that does not require additional fine-tuning, specifically leveraging a training-free paradigm using pre-trained VLLMs for table recognition. Recognizing the absence of dedicated benchmarks in this domain, we propose an evaluation benchmark based on a hierarchical design philosophy \cite{sui2024table,cheng2025survey,liu2024generative}, comprising recognition tasks for table recognition. Through extensive evaluations, we identify a critical bottleneck: low-quality input images significantly hinder the table recognition capabilities of the evaluated VLLMs. 

To overcome this limitation, we propose the Neighbor-Guided Toolchain Reasoner (NGTR) framework for effective table recognition. One of the key features of the framework is its integration of lightweight models and the strategy of retrieval-augmented generation to improve image quality and guide structured data recognition. Specifically, we propose a preprocessing toolkit with various lightweight models to enhance input image quality. For each input instance, we retrieve a similar neighbor from the training data and use the experience gained from that neighbor to guide the generation of tool invocation plans. Furthermore, we incorporate a reflection-driven tool selection module at each step to iteratively refine the table recognition output. This enables VLLMs to produce more accurate structured data.

To validate the effectiveness of the proposed NGTR framework, we conduct extensive experiments on multiple public table recognition datasets. The key observations are as follows: (1) Our NGTR framework significantly enhances the table recognition performance of naive VLLM-based approaches; (2) While VLLMs achieve competitive accuracy on specific datasets compared to traditional models, a noticeable performance gap remains in favor of traditional models. Nonetheless, we have preliminarily revealed the performance boundaries of VLLMs in several representative table recognition datasets. As is shown in Figure~\ref{motivation}, the VLLM-based table recognition approach demonstrates the capability for universal modeling. This method facilitates a paradigm shift in design objectives from a model-centric to a data-centric focus,  presenting significant potential for further exploration. We hope this work will inspire more research efforts in the future. 
In summary, the contributions of this paper are as follows:

\begin{itemize}
    \vspace{-0.02in}
    \item We conduct a systematic investigation into VLLM-based table recognition by introducing a hierarchical benchmark for evaluating their recognition capabilities.
    \item We propose the NGTR framework to address critical bottlenecks in table recognition, such as low-quality input images.
    \item We conduct extensive experiments to report the promising performance and potential of VLLMs for table recognition, along with interesting observations that highlight areas for future research.
\end{itemize}

\vspace{-0.15in}
\section{Related Work}

\textbf{Table Recognition}. Earlier table recognition (TR) methods predominantly rely on heuristic rules \cite{TR-Heu-2,TR-Heu-3} or statistical learning techniques \cite{TR-Stat-1}, these approaches rely heavily on handcrafted features or implicit rules and show limited generalization ability.  
In the era of deep learning, numerous studies have made impressive progress in handling more intricate and heterogeneous table structures. 
\textit{Top-down methods} \cite{TR-TD-1,TR-TD-2,TR-TD-3,qin2024semv3} predict table borders to infer the structure information. 
\textit{Bottom-up methods} \cite{zheng2021global,qiao2021lgpma,xing2023lore} first identify table cells with object detection models \cite{OD-1,OD-2}, and then predict the cell relations to organize the row-column structures to form the overall tables. 
These methods follow an explicit two-stage learning paradigm with relatively strong transferability and explainability, yet the risks of ambiguous contents and boundless structures may lead to unstable and incorrect prediction results.

Recently, \textit{sequence-based methods} \cite{zhong2020image,TR-Seq-3,TR-Seq-4} have been widely explored to directly generate markup sequence defining structures with specific decoders. 
Although these approaches require a massive of training data and computing resources, they have demonstrated substantial potential to unify visual-text parsing tasks \cite{wan2024omniparser}.

\noindent \textbf{Large Language Models}. 
In recent years, LLMs have demonstrated exceptional performance in tasks such as multi-task learning \cite{chen2024multi}, zero-shot learning \cite{kojima2022large}, and text generation \cite{li2024pre}. LLMs have not only broken through the limitations of traditional technologies in processing natural language text, but have also shown capabilities in reasoning \cite{wei2022chain} and planning \cite{guan2023leveraging,gou2023critic}.
Meanwhile, VLLMs combine visual and language understanding capabilities, enabling LLMs to process visual information. For multimodal understanding scenarios (\textit{e.g.}, scene text recognition \cite{STR}, visual question answering \cite{VQA}), VLLMs have been widely validated as effective \cite{Image2LLM,UReader,TextMonkey}.
With continuous progress in text-rich scenarios, some studies \cite{zheng2024multimodaltableunderstanding,zhao2024tabpedia,chen2023tablevlm} have also focused on enabling VLLMs to handle multimodal table understanding tasks.

Despite their promising success in various domains, VLLMs applied to TR remain under-evaluated and under-explored. Our study presents a comprehensive benchmark for VLLM-based TR evaluation. Subsequently, we propose a novel framework to address the bottleneck of VLLMs, thereby enhancing their capabilities in TR.

\section{Preliminary and Proposed Benchmark}

\subsection{Problem Definition}
We employ the generation paradigm of VLLMs to address the table recognition (TR) task, which is formulated as a format mapping problem from images to sequences.
Formally, given a TR dataset $\mathcal{D}=\{(I^i,H^i)\}_{i=1}^n$ with $n$ samples, we predict the corresponding structured form $H^i$ for each table image $I^i$. 
Specifically, we provide the image table $I^i$ along with a prompt $P$  as input to the VLLMs, which generates the structured data form $\hat{H}^i=\operatorname{VLLM}(P,I^i)$.

\subsection{Benchmark Evaluation Setup} \label{benchmark-setup}
This section proposes an evaluation benchmark for TR based on VLLMs, outlining the hierarchical recognition tasks to assess their performance and the evaluation setup, including the recognition tasks and VLLMs being evaluated. Details of the evaluation metrics and datasets are discussed in Section \ref{exp setup}.

\subsubsection{Recognition Task Design} \label{fine-task}
We design several hierarchical recognition tasks to conduct a more in-depth assessment of VLLMs' table recognition capability. In addition to the general table recognition task, we extend our approach by designing recognition tasks that focus on the structural composition of the table,  including the cell-level, row-level, column-level, and global table-level. This hierarchical recognition tasks design comprehensively assesses VLLM's table recognition capability at different levels. Table \ref{benchmark} presents the details of the hierarchical recognition tasks.

\textbf{Cell-level}. We evaluate the cell-level recognition capability of VLLMs within three specific recognition tasks. 
\textit{Merged cell detection} task aims to recognize cells that span multiple rows or columns in the table.
\textit{Content/index-based cell recognition} tasks evaluate the structural and content recognition capabilities of VLLMs, which are crucial for assessing whether VLLMs could perform well in fine-grained table recognition. 

\textbf{Row/Column-level}. 
We evaluate the row/column-level recognition capability of VLLMs within two specific recognition tasks.  
\textit{Row/column index-based data recognition} tasks are designed to assess whether VLLMs can accurately identify row/column elements. 

\textbf{Table-level}. 
We evaluate the overall table-level recognition capability of VLLMs. 
The \textit{visual table size detection} recognition task is designed to evaluate whether VLLMs can comprehend the global structural information and accurately determine the number of rows and columns in a table.

\begin{table}[t]
\vspace{-0.1in}
\caption{Descriptions of the proposed hierarchical recognition tasks.}
\vspace{-0.1in}
\renewcommand{\arraystretch}{1.3}
\label{benchmark}
\scalebox{0.72}{
\begin{tabular}{ccll}
\hline
\textbf{Granularity} & \textbf{Recognition Task} & \multicolumn{1}{c}{\textbf{Description}} \\ \hline
\multirow{1}{*}{Table-level}
& \begin{tabular}[c]{@{}c@{}}Visual Table Size\\ Detection\end{tabular} & Get the number of rows and columns.
\\ \hline
Row-level & \begin{tabular}[c]{@{}c@{}}Row Index-based\\ Data Recognition\end{tabular} & Get the content list of a specific row. \\ \hline
Column-level & \begin{tabular}[c]{@{}c@{}}Column Index-based\\ Data Recognition\end{tabular} & Get the content list of a specific column. \\ \hline
\multirow{5}{*}{Cell-level} & \begin{tabular}[c]{@{}c@{}}Merged Cell\\ Detection\end{tabular} & Get contents of all merged cells. \\ \cline{2-3}
& \begin{tabular}[c]{@{}c@{}}Content-based\\ Cell recognition\end{tabular} & Get the location of specific cell content. \\ \cline{2-3}
& \begin{tabular}[c]{@{}c@{}}Index-based\\ Cell Recognition\end{tabular} & Get the cell content of specific location. \\ \hline
\end{tabular}
}
\vspace{-0.15in}
\end{table}
\vspace{-0.05in}
\subsubsection{Baselines}
In this study, we evaluate the performance of six VLLMs. For open-source VLLMs, we select Phi-3.5 (Phi) and Llama-3.2-90B (Llama) for evaluation. For closed-source VLLMs, we evaluat GPT-4o-mini (GPT-mini), Qwen-VL-Max (Qwen), GPT-4o (GPT) and Gemini-1.5-Pro  (Gemini).

\begin{figure}[t] \centering
    \includegraphics[width=0.42\textwidth]{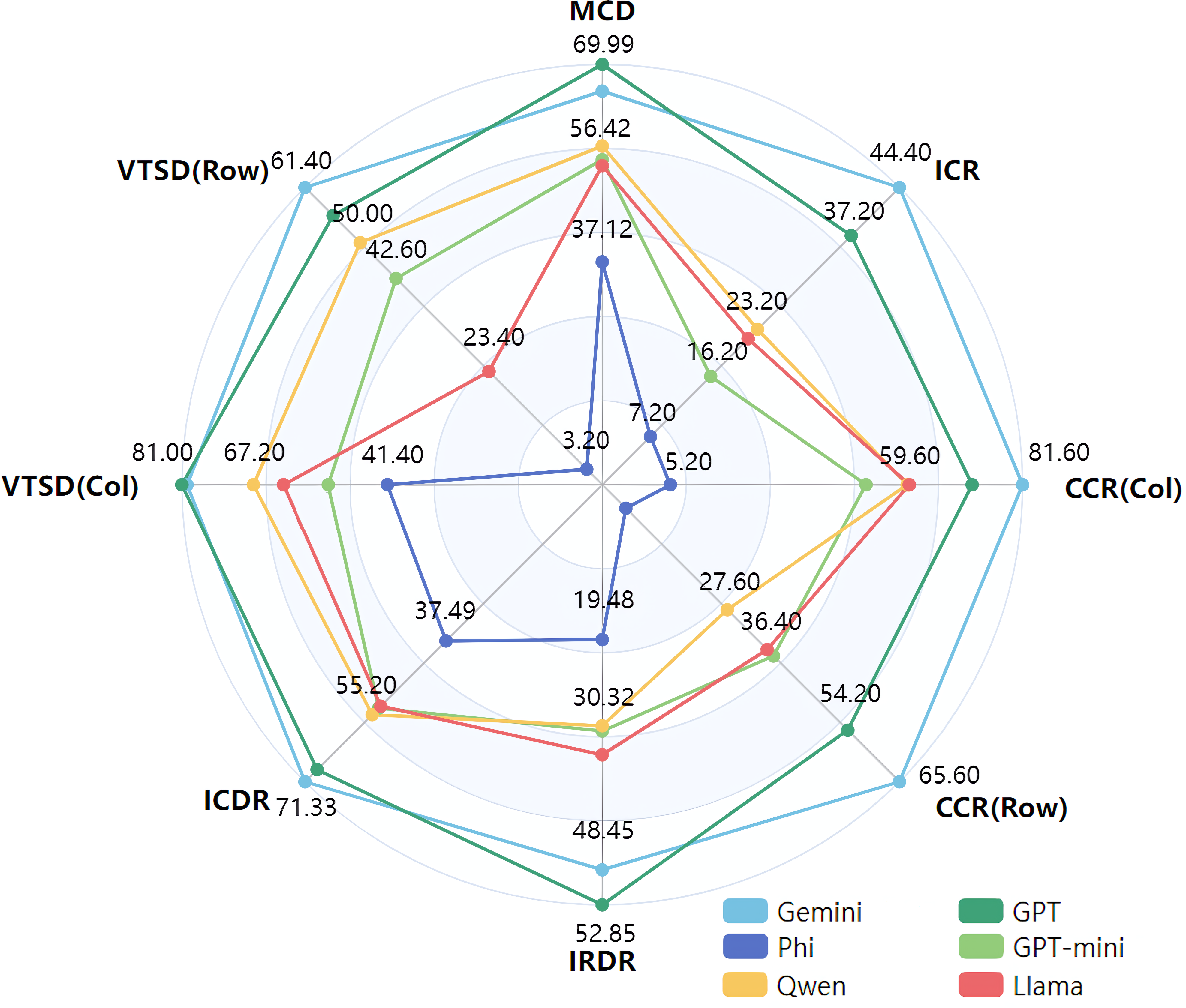}
    \vspace{-0.1in}
    \caption{Experimental results of VLLMs for the proposed hierarchical tasks. The tasks evaluated include the following: merged cell detection (MCD), content/index-based cell recognition (CCR, ICR), index-based row/column data recognition (IRDR, ICDR), and visual table size detection (VTSD).}
    \label{fig:subTask}
    \vspace{-0.2in}
\end{figure}

\subsection{Benchmark Evaluation Results} \label{fine-grained experimental results}
\begin{figure}[h] \centering
	\vspace{-0.12in}
    \includegraphics[width=0.22\textwidth]{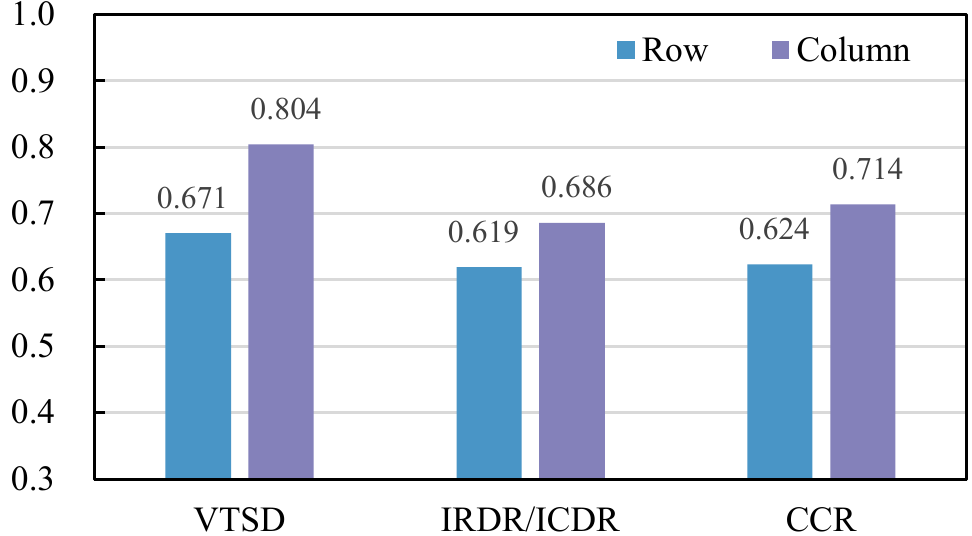}
    \includegraphics[width=0.22\textwidth]{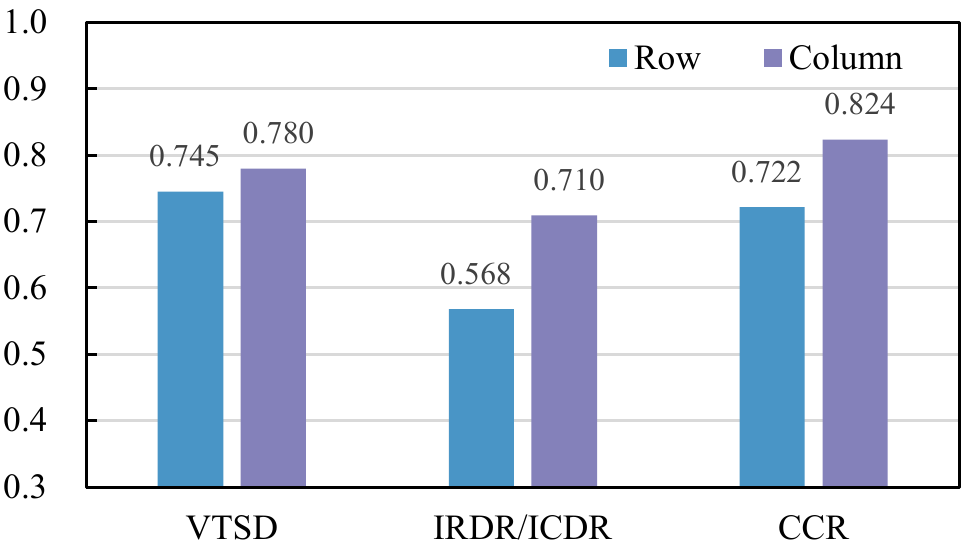}
  	\vspace{-0.1in} 
    \caption{Row-Column Sensitivity Analysis of VLLMs on Hierarchical Tasks with Gemini (Left) and GPT (Right).} 
    \label{fig:rc}
    \vspace{-0.1in}
\end{figure}

Evaluation results of hierarchical recognition tasks are presented in Figure \ref{fig:subTask}.
Among all the VLLMs we selected, GPT and Gemini demonstrate the strongest performance, consistently outperforming the other VLLMs. Furthermore, the open-source Llama demonstrates a significant performance gap compared to the closed-source VLLMs. We give some highlights associated with the benchmark results as follows:

\textbf{Row-column Sensitivity Analysis.} We found that all models show inconsistent performance between row/column-related tasks. To mitigate the influence of uneven distributions of rows and columns in the original data leading to varying difficulty levels, we further refine the experimental results by selecting samples where the difference between the number of rows and columns does not exceed three.
The results are presented in Figure \ref{fig:rc}, showing that the accuracy for the column-related tasks is higher than for the row-related tasks, suggesting that VLLMs prefer to process column-structured data. 
This phenomenon likely occurs because columns usually exhibit high diversity, representing different attributes, whereas rows exhibit high similarity, typically corresponding to different entities. As a result, VLLMs are skilled in handling column-related tasks, where attribute information facilitates more accurate recognition.

\textbf{Image Quality Analysis.} 
We conduct in-depth experiments and analysis on the TR task. Detailed experimental results are presented in Section \ref{exp}. The results demonstrate that the VLLMs perform relatively well on the SciTSR dataset with higher image quality using simple prompts. However, the performance gap is considerably more significant when processing the PubTabNet dataset with lower image quality, especially compared to traditional models. This phenomenon indicates that the quality of the input image is a key bottleneck limiting the performance of VLLMs. Section \ref{Bottleneck res} provides a more in-depth analysis of this bottleneck.

\begin{table}[t]
\vspace{-0.1in}
	\caption{Specific description of the challenges and scenarios in low-quality image inputs.}
	\vspace{-0.05in}
    \renewcommand{\arraystretch}{1.15}
	\label{tabl:bottle}
	\scalebox{0.8}{
\begin{tabular}{ccll}
\hline
\textbf{Challenge} & \textbf{Scenario} & \multicolumn{1}{c}{\textbf{Description}} \\ \hline
\multirow{5}{*}{\begin{tabular}[l]{@{}c@{}}Visual\\ Conditions\end{tabular}} & Blur & 
\begin{tabular}[l]{@{}l@{}} The image is out of focus, with details\\ appearing smeared or indistinct.\end{tabular} \\ \cline{2-3}
 & Underexposure & \begin{tabular}[l]{@{}l@{}}The image is too dark, which may \\cause the content to be unclear.\end{tabular} \\ \cline{2-3}
 & Overexposure & \begin{tabular}[l]{@{}l@{}}The image is overly bright, losing\\ detail in some regions.\end{tabular} \\ \hline
\multirow{5}{*}{\begin{tabular}[l]{@{}c@{}}Table \\ Border\\ Quality\end{tabular}} & \begin{tabular}[c]{@{}c@{}}Unclear\\ Borders\end{tabular} & \begin{tabular}[l]{@{}l@{}}The image's table borders are faint,\\ blending into the background.\end{tabular} \\ \cline{2-3}
 & \begin{tabular}[c]{@{}c@{}}Missing \\ Borders \end{tabular} & \begin{tabular}[l]{@{}l@{}}The table is without expected \\ borders or separators.\end{tabular} \\
 \cline{2-3}
 & \begin{tabular}[l]{@{}c@{}}Thickened\\ Borders\end{tabular} & The table borders are thickened. \\ \hline
\multirow{2}{*}{\begin{tabular}[l]{@{}c@{}}Geometric\\ Deformation\end{tabular}} & Tilt 20° &  The image is tilted at 20° angle.\\ 
\cline{2-3}
 & Tilt 40° & The image is tilted at 40° angle. \\ \cline{1-3}
\end{tabular}
}
\vspace{-0.14in}
\end{table}

\vspace{-0.06in}
\subsection{Bottleneck Analysis} \label{Bottleneck res}
We further investigate the performance of VLLMs under varying image quality conditions and assess their visual robustness to these conditions through empirical analysis.

\vspace{-0.04in}
\subsubsection{Experimental Setup} \label{visual-task}
To comprehensively evaluate the visual robustness of VLLM, we focus on three distinct visual challenges: the image quality challenge, the table border quality challenge, and the geometric deformation challenge. 
Details of these challenges are provided in Table \ref{tabl:bottle}.

\textbf{Visual Conditions}. Visual conditions is a key factor affecting the accuracy of VLLMs in table recognition. 
To assess it, we systematically analyze the performance of VLLMs under various visual conditions across three scenarios: \textit{blur}, \textit{underexposure}, and \textit{overexposure}. 
These analyses demonstrate the robustness of VLLMs in handling impaired visual conditions.

\textbf{Table Border Quality}. Table borders indicate structural information in table elements. We evaluate the impact of border visibility and completeness on table recognition performance by considering scenarios including \textit{unclear table borders} and \textit{missing borders}. Additionally, we explore the effect of border changes in the table by \textit{thickened table borders}.

\textbf{Geometric Deformation}. Geometric deformation caused by viewing angles or operations can disrupt the geometric consistency of tables.
We evaluated the robustness of VLLMs against geometric deformations by testing them \textit{under tilt conditions of 20°} and \textit{excessive tilt conditions of 40°}.

\vspace{-0.04in}
\subsubsection{Discussion and Analysis}
\begin{figure}[t] 
        \vspace{-0.1in}
	\centering
	\includegraphics[width=0.48\textwidth]{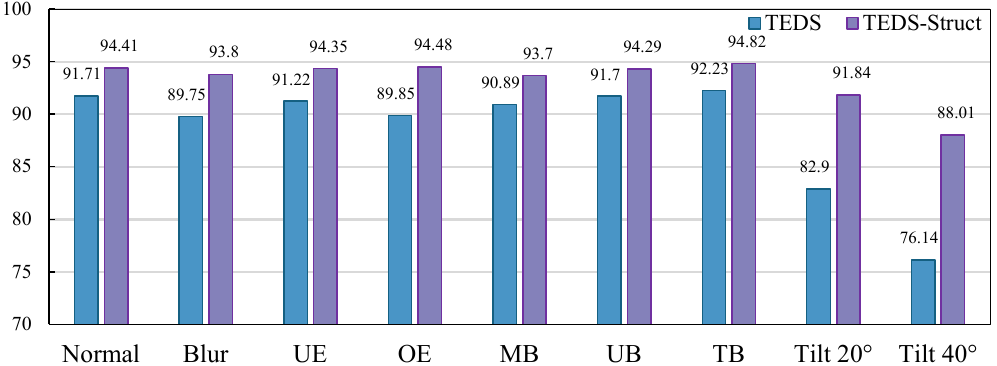}
	\vspace{-0.2in}
	\caption{Evaluation results of bottleneck scenarios: abbreviations UE (Underexposure), OE (Overexposure), MB (Missing Borders), UB (Unclear Borders), TB (Thickened Borders).} 
	\label{fig:errorTask}
	\vspace{-0.14in}
\end{figure}

Figure \ref{fig:errorTask} shows the detailed experimental results. We observe that blurring and overexposure can lead to loss of text information in table images and inaccurate fine-grained text recognition, thus negatively impacting the performance of table content recognition. Moreover, VLLM demonstrates limited efficacy when processing skewed tables, resulting in a substantial decrease in accuracy under such conditions.

Although table borders are often considered essential for conveying structured information, fading or removing these borders has minimal impact on the performance of VLLMs. This result suggests that VLLMs do not heavily rely on borders. VLLMs only pay slight attention to the structural information the borders provide when thickening.

\begin{figure*}[ht]
    \vspace{-0.15in}
    \centering
    \includegraphics[width=\textwidth]{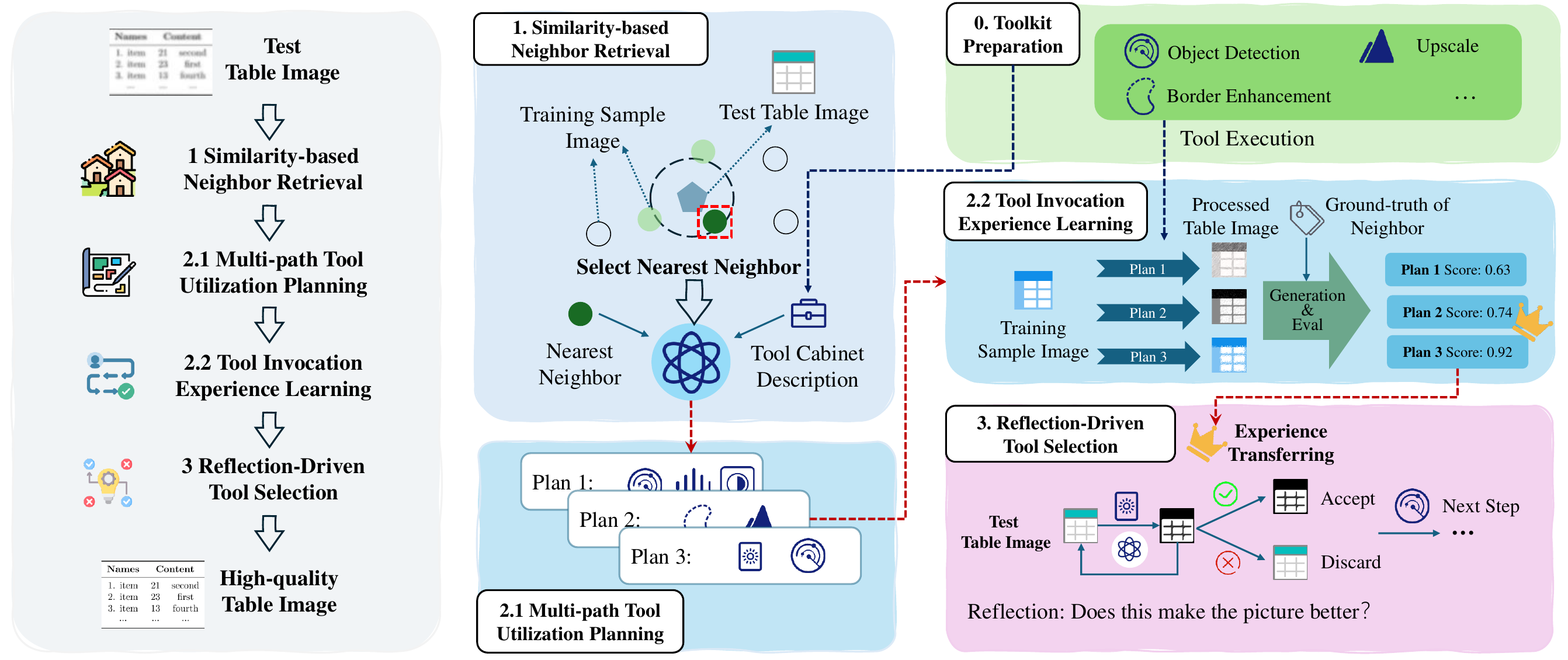}
    \vspace{-0.2in}
    \caption{Illustration of a pipeline for table image preprocessing leveraging a toolkit of lightweight vision models.}
    \label{main figure}
    \vspace{-0.15in}
\end{figure*}

\section{Neighbor-Guided Toolchain Reasoner}
Through our in-depth analysis of VLLMs' performance on the benchmark, we have identified that 
improving input image quality is essential for enhancing VLLMs' capability to recognize and interpret structured image data more effectively. To address this, we propose the NGTR framework.

\subsection{Framework Overview}

NGTR enhances input image quality by applying various tool combinations tailored to low resolution, overexposure, and noise interference. As shown in Figure \ref{main figure}, we design a similarity-based neighbor retrieval module to select a suitable combination of tools. Subsequently, the tool invocation experience learning module executes each plan and generates the corresponding structured data to evaluate the effectiveness of different plans. Finally, we propose a reflection-driven tool selection module to integrate iterative tool invocation and dynamic feedback to refine the processing flow. The optimized image is then input into the VLLMs, which utilize their powerful reasoning capabilities to generate structured data.

\vspace{-0.04in}
\subsection{Toolkit Preparation}

Inspired by the conclusions of the Bottleneck Analysis (Section \ref{Bottleneck res}), we employ five distinct tools to address various scenarios and potential issues that may arise in the table recognition task. These tools are shown in Table \ref{tab:tools}. By combining different tools, NGTR effectively addresses various challenging situations. For example, when the table border is faint, the VLLMs can invoke the border enhancement tool to strengthen the structural information by thickening the table border. Similarly, when the table occupies only a portion of the image, the VLLMs can invoke the table detection and cropping tool to identify and crop the relevant table area, thereby reducing noise interference.

\subsection{Similarity-based Neighbor Retrieval } \label{section-sim-neighbor}
Neighbor retrieval methods enable VLLMs to retrieve similar neighbor samples, providing richer contextual information. In the NGTR framework, we hypothesize that images with similar features exhibit similar results after being processed by the same image preprocessing toolchain. Consequently, the processing results of neighbor samples could guide the selection of a potentially optimal toolchain for test samples. We first retrieve images that are similar to the target task images from a sample set. Subsequently, we employ a prompting template to leverage the VLLM's planning capabilities to generate tool invocation plans. The retrieval process can be formally described as follows:
\begin{equation}
	\setlength{\abovedisplayskip}{2pt}
	\setlength{\belowdisplayskip}{2pt}
\text{Retrieval}(I^{\text{test}}, \mathcal{D}^{\prime}, f) = \arg\max \left[ f\left(I^{\text{test}}, I^i\right) \right]_{i=1}^{|\mathcal{D}^{\prime}|},
\end{equation}
where \(I^{\text{test}}\) represents an image from the test set, \(\mathcal{D}^{\prime}\) denotes a subset of the training dataset, and \(f\) is the similarity measurement function. In this paper, we combine the ORB (Oriented FAST and Rotated BRIEF) algorithm with the Hamming distance as \(f\) to measure the similarity between images. 
Then, we guide the VLLMs to generate multiple tool invocation plans for the image. The generation process can be formally expressed as follows:
\begin{equation}
	\setlength{\abovedisplayskip}{3pt}
	\setlength{\belowdisplayskip}{3pt}
\text{VLLM}(\mathcal{T}, \mathcal{N}(I^{\text{test}})) \rightarrow \{p_1, p_2, \ldots, p_n\},
\end{equation}
where \(\mathcal{T}\) represents the description information set of all available image preprocessing tools, including their functions, applicable scenarios, invocation identifiers, and other relevant details; \(\mathcal{N}(I^{\text{test}})\) denotes the neighbor image samples of the test sample \(I^{\text{test}}\), along with their associated features, retrieved from the training set; and \(\{p_1, p_2, \ldots, p_n\}\) represents the generated candidate set of plans, which are then used to select an appropriate tool invocation plan.

\begin{table}[t]\centering
	\caption{Specific descriptions of built tools in the toolkit.}
	\label{tab:tools}
	\vspace{-0.05in}
	\scalebox{0.76}{
		\large
		\renewcommand{\arraystretch}{1.05} 
		\begin{tabular}{>{\centering\arraybackslash}m{0.10\textwidth} >{\raggedright\arraybackslash}m{0.45\textwidth}}
			\toprule
			\textbf{Tool} & \textbf{Descriptions} \\
			\midrule
			Border Enhancement & The border enhancement tool improves the legibility of tables and their structures by thickening the border lines in the image. This process enhances the structural information features of tabular data. \\
			\midrule
			Image Upscaling & Image upscaling optimize image resolution to improve visual quality. This technique is commonly employed to repair and enhance blurry images. \\
			\midrule
			Noise Reduction & The noise reduction tool enhances image quality by adjusting brightness and contrast to mitigate noise interference and underexposure issues. \\
			\midrule
			Binarization & This tool converts images to black and white, highlighting key features for easier extraction. \\
			\midrule
			Detection and Cropping & This tool identifies table regions within an image and crops them into independent segments. It is particularly suitable for processing images of tables embedded in complex backgrounds. \\
			\bottomrule
		\end{tabular}
	}
   \vspace{-0.2in}
\end{table}

\vspace{-0.05in}
\subsection{Tool Invocation Experience Learning}

In this module, we follow a sequential workflow to evaluate the multiple tool invocation plans. First, we execute each tool invocation plan generated by the previous module to obtain multiple processed images:
\begin{equation}
	\setlength{\abovedisplayskip}{3pt}
	\setlength{\belowdisplayskip}{3pt}
I_{p_i} = f_{p_i}(I), \quad i \in \{1, 2, \ldots, n\},
\end{equation}
where \( f_{p_i} \) denotes the image preprocessing tools.
Next, we employ a prompt template to guide the VLLMs in generating a markup sequence.
Subsequently, we evaluate the prediction results based on the example labels. The evaluation process employs the tree edit distance-based similarity (TEDS) metrics to quantify the accuracy of the VLLMs output.
By following this process, we calculate a quantitative score for each toolchain, enabling the selection of a suitable plan.

\vspace{-0.08in}

\begin{table*}[t]
	\centering
	\definecolor{customgreen}{RGB}{230, 248, 224} 
	\definecolor{customblue}{RGB}{212, 230, 241} 
	\caption{Performance comparison of methods on the SciTSR, PubTabNet, and WTW datasets. "-" indicates the method's lack of results (specific reasons are provided in the implementation details). Best scores in the lightweight OCR model category are highlighted in \colorbox{customblue}{blue}, while best scores in the prompt tuning in VLLMs category are highlighted in \colorbox{customgreen}{green}.}
	\vspace{-0.1in}
	\renewcommand{\arraystretch}{1.25}
	\scalebox{0.75}{
	\begin{tabular}{cc|ccc|cccccccccc}
		\toprule
		\multirow{3}[4]{*}{\textbf{Dataset}} & \multirow{3}[4]{*}{\textbf{Metrics}} & \multicolumn{3}{c|}{\textbf{Lightweight OCR Model}} & \multicolumn{10}{c}{\textbf{Prompt Tuning in VLLMs}} \\
		\cmidrule{3-15} 
		& & \multirow{2}[2]{*}{EDD} & \multirow{2}[2]{*}{LGPMA} & \multirow{2}[2]{*}{LORE} & \multirow{2}[2]{*}{Phi} & \multirow{2}[2]{*}{GPT-mini} & \multirow{2}[2]{*}{Qwen} & \multirow{2}[2]{*}{Llama} & \multicolumn{3}{c}{Gemini} & \multicolumn{3}{c}{GPT-4o} \\
		\cmidrule{10-12}
		\cmidrule{13-15}
		          &       &       &       &       &       &       &       &       & direct & NGTR & $\Delta$ & direct & NGTR & $\Delta$ \\
		\midrule
		\multirow{2}[1]{*}{SciTSR} & TEDS  & -     & \cellcolor{customblue}\textbf{95.08}  & -     & 66.18  & 87.18  & 89.40  & 87.24  & 90.15  & 91.07 & +0.92 & 90.70  & \cellcolor{customgreen}\textbf{92.58} & +1.88  \\
		& TEDS-Struct & -     & 96.24  & \cellcolor{customblue}\textbf{97.22}  & 71.56  & 92.03  & 93.06  & 92.31  & 93.73  & 95.09 & +1.36  & 94.20  & \cellcolor{customgreen}\textbf{95.78} & +1.58  \\
		\midrule
		\multirow{2}[0]{*}{PubTabNet} & TEDS  & 89.67  & \cellcolor{customblue}\textbf{94.63}  & -     & 49.92  & 58.68  & 52.53  & 79.04  & 81.00  & 84.80 & +3.80  & 74.46  & \cellcolor{customgreen}\textbf{85.03} & +10.57  \\
		& TEDS-Struct & -  & 96.70  & \cellcolor{customblue}\textbf{96.94}  & 57.65  & 73.00  & 63.90  & 87.64  & 85.28  & 89.30 & +4.02 & 84.91  & \cellcolor{customgreen}\textbf{92.31} & +7.40  \\
		\midrule
		WTW & TEDS-Struct & -     & -     & \cellcolor{customblue}\textbf{93.86}  & -     & 31.72  & -  & 32.87  & 42.62  & 44.68 & +2.06  & 40.01  & \cellcolor{customgreen}\textbf{52.03} & +12.02 \\
		\bottomrule
	\end{tabular}
}
	\label{tab:mainTable}
    \vspace{-0.15in}
\end{table*}

\subsection{Reflection-driven Tool Selection} 
\label{section-reflection}

Although the tool invocation experience learning module provides a high-quality plan, mindlessly applying the tool invocation plan to new samples may result in the loss of critical information in the image, thereby affecting the accuracy of the final result. To address this, we introduce the reflection-driven tool selection module during the execution phase to refine the processing flow, reduce information loss, and thereby improve recognition accuracy. The formalized expression of the reflection module is as follows:

Let \( I^{(t - 1)} \) denote the image before the \( t \)-th operation and \( I^{(t)} \) denote the image after the \( t \)-th operation.  
The VLLMs computes \( \gamma^{(t)} \) to determine whether to accept the operation: 
\begin{equation}
\setlength{\abovedisplayskip}{3pt}
\setlength{\belowdisplayskip}{3pt}
\gamma^{(t)} = \text{reflect}(I^{(t-1)}, I^{(t)}),
\end{equation}
where \( \gamma^{(t)} \) is a binary decision indicating the quality change between the before and after images. The function \( \text{reflect}(\cdot) \) evaluates the difference in quality. If \( \gamma^{(t)} = 1 \), the operation is considered successful; otherwise, if \( \gamma^{(t)} = 0 \), the operation is rejected, and the process proceeds to the next step.

This step-by-step module enhances the interaction between the VLLMs and the target image, ensuring the accuracy of the final task outcome. More importantly, introducing this module enables downstream researchers and developers to flexibly customize and expand the toolkit without worrying about the impact of poorly performing expanded tools on the final results, thereby significantly improving the versatility and transferability of our framework. In the last step, we design a simple prompt template to instruct VLLM to generate a markup sequence and obtain the result of table recognition.

\vspace{-0.1in}
\section{Experiments} \label{exp}

\subsection{Experimental Setup} \label{exp setup}
\vspace{-0.02in}
\subsubsection{Datasets}
In this study, we utilize three widely-used table recognition datasets: SciTSR \cite{chi2019complicated}, PubTabNet \cite{zhong2020image}, and WTW \cite{long2021parsing}, each offering unique characteristics and challenges.
SciTSR is a dataset comprising tables extracted from the scientific literature, and the image quality in this dataset is relatively high.
In contrast, the image resolution of PubTabNet is 72 pixels per inch, and its overall image quality is relatively low. WTW contains images collected from the wild, introducing a variety of extreme cases, such as tilt, blur, and table curvature.
These datasets encompass diverse table types and various unique visual challenges, providing a robust foundation for benchmarking.


\vspace{-0.02in}
\subsubsection{Baselines}
We select six Vision Large Language Models (VLLMs), including Phi\footnote{\url{https://azure.microsoft.com/en-us/products/phi/}}, Llama\footnote{\url{https://www.llama.com/}}, GPT-mini, Qwen\footnote{\url{https://qwenlm.github.io/blog/qwen-vl/}}, GPT\footnote{\url{https://openai.com/index/hello-gpt-4o/}}, and Gemini\footnote{\url{https://deepmind.google/technologies/gemini/pro/}}, as baseline models for comparison. Additionally, we select three representative deep learning-based methods as baselines for comparison: EDD \cite{zhong2020image} based on sequence modeling, LGPMA \cite{qiao2021lgpma} based on cell bounding box detection, and LORE \cite{xing2023lore} based on cell point center detection.

\vspace{-0.02in}
\subsubsection{Evaluation Metrics} \label{Evaluation Metrics}

As for the evaluation metrics of TR, we use a similarity metric based on Tree-Edit Distance (TEDS) \cite{zhong2020image} and the TEDS-Struct metric. We employ two evaluation metrics for the hierarchical tasks described in Section \ref{fine-task}: accuracy (ACC) and micro-averaged F1 score (F1-score). Specifically, ACC is used to evaluate cell-level tasks (excluding merged cell detection) and table-level tasks; the F1-score is utilized for row- and column-level tasks and merged cell detection.

\vspace{-0.02in}
\subsubsection{Implementation Details} \label{Implementation details}
For the PubTabNet dataset, We randomly select 1,500 images from the validation set. For the SciTSR and WTW datasets, we use their complete test sets for evaluation. Since the WTW dataset does not provide content information for table recognition, we do not report its TEDS scores.

For LORE, since it is mainly aimed at table structure recognition but not table content recognition, we only report its performance scores for table structure recognition. 
As for EDD, since its model training requires a large amount of end-to-end annotated data, and SciTSR and WTW lack corresponding labeled data, its performance on these datasets has not been evaluated.
LGPMA depends on table content for training, but since WTW lacks content labels, its performance on this dataset was not assessed.

\subsection{Main Results Analysis}
Tables \ref{tab:mainTable} show our benchmark results on the table recognition and table structure recognition tasks. Based on the experimental results, we draw the following insights:
\vspace{-0.04in}
\subsubsection{Performance Analysis of NGTR}
As shown in Table \ref{tab:mainTable}, the experimental results compare our NGTR framework with baseline methods. The main results show that our framework achieves significant performance improvements on the PubTabNet dataset, mainly attributed to our framework's enhanced VLLMs robustness when dealing with low-quality inputs. On the SciTSR dataset, our framework also outperforms all VLLMs baselines, further verifying our framework's effectiveness.

In the WTW dataset, our method and all VLLM-based baselines demonstrate suboptimal performance compared to traditional lightweight baseline methods. However, our approach significantly enhances the effectiveness of these models, suggesting that our framework imparts a degree of robustness in complex environments. Despite this improvement, it also highlights persistent challenges that remain difficult to address. Section \ref{wtw_challenge} presents a more detailed analysis of performance on the WTW dataset.

\vspace{-0.05in}
\subsubsection{Open-source and Closed-source VLLMs} The advanced open-source VLLMs demonstrate capabilities comparable to closed-source VLLMs in this task. As a representative of open-source models, Llama exhibits outstanding performance, particularly in the table structure recognition task (TEDS-Struct) on the PubTabNet dataset. Llama achieves a relatively better result, surpassing GPT by 2.73 points and Gemini by 2.36 points. These results confirm the potential of open-source VLLMs for further research.


\vspace{-0.02in}
\subsubsection{WTW Dataset: A Challenge for VLLMs} \label{wtw_challenge}

Table \ref{tab:mainTable} presents the experimental results on the WTW dataset. These results indicate that prompt-tuning based VLLMs methods still exhibit significant gaps compared to traditional lightweight OCR methods, highlighting the challenges VLLMs face when processing datasets with wild scenarios. Further analysis of the outputs suggests that performance declines notably when VLLMs process tables with numerous empty cells, unevenly distributed text, and skewed, rotated, or densely packed text.
Further analysis of the outputs suggests that VLLMs tend to ignore cells lacking semantic content. While this behavior helps avoid processing irrelevant data, it also limits their ability to effectively capture the structural information of tables with many blank cells.

\begin{table}[t]
	\centering
    \caption{Ablation study results on key components of the framework.  "EXP" denotes the tool invocation experience learning module, "REF" represents the reflection-driven tool selection module.
    }
    \vspace{-0.06in}
    \label{tab:Ablation}
    \scalebox{0.7}{
    \large
    \renewcommand{\arraystretch}{1}
    \begin{tabular}{ccccc}
        \toprule
        \multirow{2}{*}{Method} & \multicolumn{2}{c}{SciTSR} & \multicolumn{2}{c}{PubTabNet} \\
        \cmidrule[0.5pt](lr){2-3} \cmidrule[0.5pt](lr){4-5}
         & TEDS & TEDS-Struct & TEDS & TEDS-Struct \\
        \midrule
        NGTR & 92.56 & 95.43 & 85.03 & 92.31 \\
        w/o EXP & 90.33 & 93.68 & 80.57 & 88.40 \\
        w/o REF & 91.53 & 94.77 & 82.08 & 91.85 \\
        \bottomrule
    \end{tabular}
    }
    \vspace{-0.13in}
\end{table}
\vspace{-0.04in}
\subsection{Ablation Study w.r.t Key Components}

\textbf{Effectiveness of Tool Invocation Experience Learning.} 
We performed an ablation study by removing the tool invocation experience learning module. In this experiment, we only led the VLLMs to generate a tool invocation plan and applied it directly to the test samples. As shown in Table \ref{tab:Ablation}, while the VLLMs could generate a valid tool invocation plan, the lack of effective validation of the generated plan led to a significant performance decline. This further validates the critical role of the tool invocation experience learning module in improving the NGTR framework performance.

\vspace{0.04in}
\noindent \textbf{Effectiveness of Reflection-driven Tool Selection.} 
We performed an ablation study by removing the reflection-driven tool selection module. We apply all the tools to the task images in a single pass. The results are presented in Table \ref{tab:Ablation}, without the reflection module for stepwise backtracking validation, VLLMs cannot effectively supervise the processing procedure, which may lead to the incorporation of unsuitable tools for the current sample, thereby affecting performance. 

\subsection{Tool Invocation Analysis}

We calculated the tool usage rates for each test dataset to investigate further the NGTR framework's preferences in tool invocation (i.e., the proportion of samples that invoke a particular tool among all samples where tools are invoked). The results are presented in Table  \ref{tab:toolAnalisisTable}. On the PubTabNet dataset, the NGTR framework exhibits a marked preference for the image upscaling tool. This preference is consistent with this dataset's overall lower image quality characteristic. The detection and cropping tool has the highest invocation rate on the WTW dataset, which aligns with the dataset's characteristics in the wild scenarios. In such images, tables often occupy only a portion of the frame or appear displaced, making this tool effective in locating and extracting the target areas.
In summary, the NGTR framework adaptively selects appropriate image processing tools based on the characteristics of different datasets, thereby fully showcasing its flexible planning capabilities in cross-dataset scenarios.

\begin{table}[t]\centering
	\caption{Tool invocation preference analysis. Values show the percentage of samples using each tool, with bold values marking the dataset where each tool is most frequently used.}
	\vspace{-0.06in}
	\label{tab:toolAnalisisTable}
	\resizebox{0.48\textwidth}{!}{
		\large
		\renewcommand{\arraystretch}{1.25}
		\begin{tabular}{*{10}{c}}
			\toprule
			Dataset & \begin{tabular}[c]{@{}c@{}}Image\\ Upscaling\end{tabular} & \begin{tabular}[c]{@{}c@{}}Border\\ Enhancement\end{tabular} & Binarization & \begin{tabular}[c]{@{}c@{}}Noise\\ Reduction\end{tabular} & \begin{tabular}[c]{@{}c@{}}Detection and\\ Cropping\end{tabular} \\ \midrule
			PubTabNet & \textbf{93.71} & 76.42 & 4.40 & 15.64 & 64.39 \\
			SciTSR    & 77.97 & \textbf{90.79} & 3.70 & \textbf{20.64} & 54.23 \\
			WTW       & 79.81 & 86.86 & \textbf{6.65} & 15.46 & \textbf{82.93} \\
			\bottomrule
		\end{tabular}
	}
	\vspace{-0.1in}
\end{table}

\begin{figure}[h] 
        \vspace{-0.06in} 
	\centering
	\includegraphics[width=0.22\textwidth]{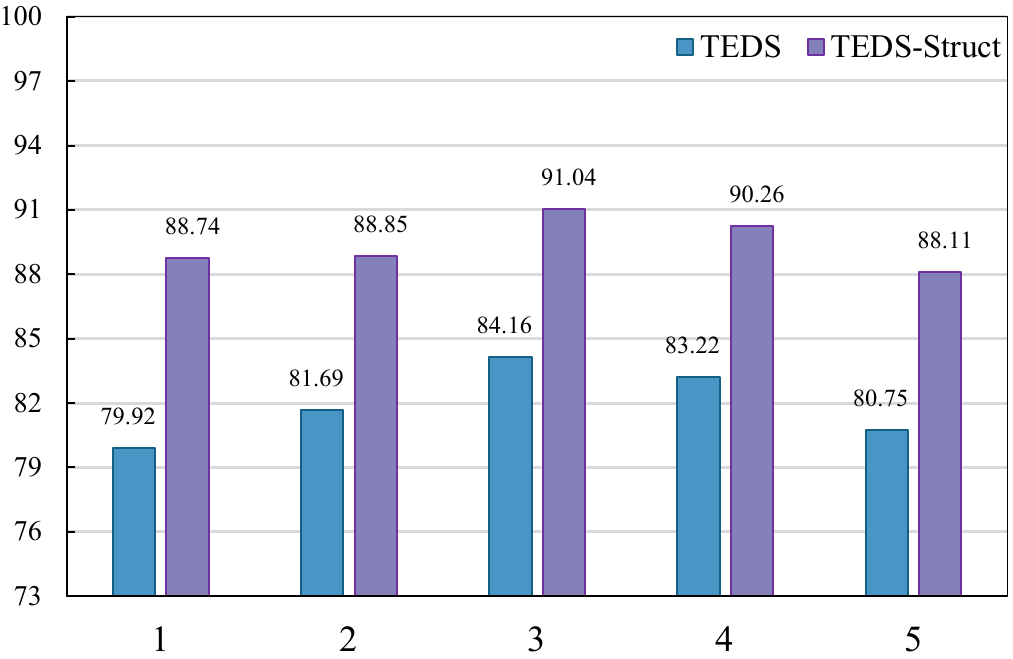}
	\includegraphics[width=0.22\textwidth]{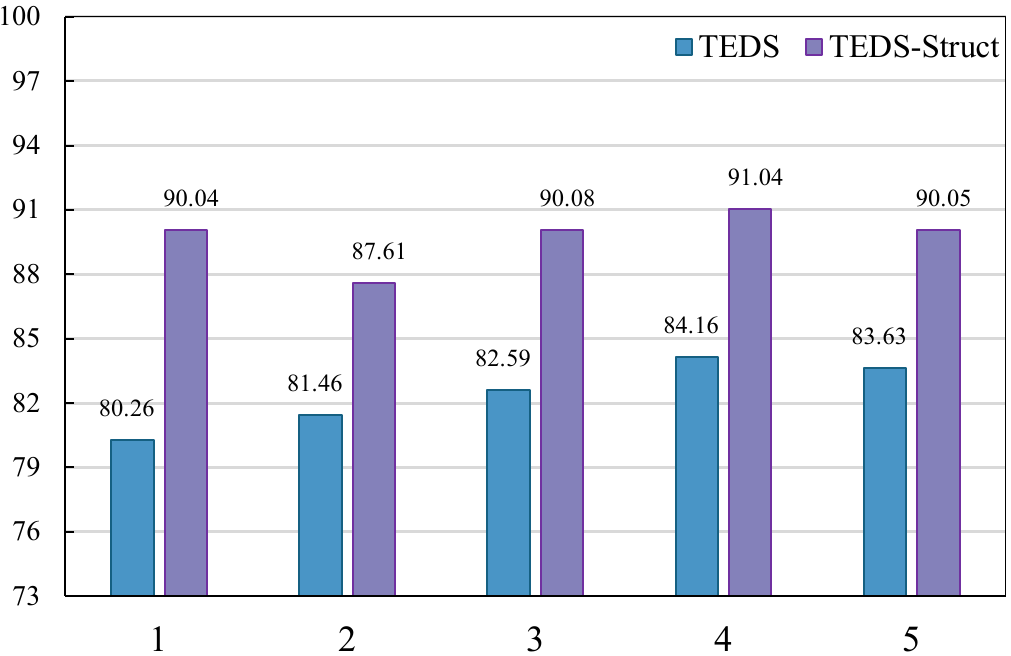}
	\vspace{-0.06in} 
	\caption{The impact of the number of tools (shown on the left) and the number of multi-paths (shown on the right) on performance.} 
	\label{fig:pathNtoolN}
	\vspace{-0.16in}
\end{figure}

\vspace{-0.06in}
\subsection{Hyperparameter Sensitivity Analysis}

The NGTR framework contains two core parameters: the maximum length of the toolchain execution plan \( L \) and the number of plans generated each time \( N \). As shown in Figure \ref{fig:pathNtoolN}, a moderate toolchain length achieves an adequate balance between complexity and performance, as excessive toolchain length increases combinatorial complexity and limits processing performance, thereby affecting the framework's ability to generate high-quality solutions. Similarly, generating a moderate number of execution plans effectively balances solution quality and generation efficiency, whereas generating too few or too many plans slightly reduces performance. Therefore, a moderate toolchain length and number of execution plans can balance complexity and performance well, providing valuable guidance for the tool invocation.

\vspace{-0.02in}
\section{Conclusion and Limitation}
This paper addressed the under-explored challenge of table recognition using VLLMs in a training-free paradigm. We proposed the NGTR framework, which enhanced input image quality through lightweight models and neighbor-guided tool invocation strategies. Extensive experiments demonstrated that NGTR significantly improved VLLM-based table recognition performance. This work not only established a benchmark for table recognition but also highlighted the potential of VLLMs in advancing table understanding, paving the way for future research and applications.

\noindent \textbf{Limitation}. Despite the strengths of our framework, we acknowledge several limitations that warrant further investigation. Firstly, its performance depends on the underlying toolkit. Secondly, when the available set of neighbor candidates does not sufficiently cover a wide range of scenarios, selecting an inappropriate neighbor may lead to suboptimal performance. Nevertheless, we believe the NGTR framework demonstrates strong generalizability, serving as a versatile approach for tool invocation for various domains.

\clearpage

\section*{Acknowledgements}
This research was supported by grants from the grants of Provincial Natural Science Foundation of Anhui Province (No.2408085QF193), USTC Research Funds of the Double First-Class Initiative (No. YD2150002501), the National Key Research and Development Program of China (Grant No. 2024YFC3308200), the National Natural Science Foundation of China (62337001), the Key Technologies R \& D Program of Anhui Province (No. 202423k09020039) and the Fundamental Research Funds for the Central Universities (No. WK2150110032).
\bibliographystyle{named}
\bibliography{ijcai25}












\clearpage

\appendix

\begin{flushleft}
	\huge\textbf{Appendix}
\end{flushleft}

\vspace{-0.04in}
\section{Logical to Markup Sequence Conversion}
Since the baseline methods (including LORE and LGPMA) output logical structures representing cell information, they cannot be directly used for the calculation of TEDS metrics. To address this, we use the following pseudocode to convert the logical structure into markup sequence format.

We divide the entire conversion process into two stages. As shown in Algorithm \ref{algStage1}, the first stage involves a preliminary preprocessing of the logical location information, associating the logical positions with the corresponding cell content and storing them in a tabular data matrix.

\vspace{-0.08in}
\begin{algorithm}
	\raggedright
	\small
	\caption{From Logical Location to Tabular Data Matrix}
	\textbf{Input:} cells = \{ $ C_1, C_2 \dots, C_K $ \}\\
	\textbf{Output:} table
	\begin{algorithmic}[1]
		\State max\_row $\gets$ maximum value of 'end\_row' in cells
		\label{algStage1}
		\State max\_col $\gets$ maximum value of 'end\_col' in cells
		\State Initialize table as a array with dimensions (max\_row, max\_col)
		\For{cell \textbf{in} cells}
		\State start\_row, end\_row, start\_col, end\_col, content $\gets$ cell
		\State rowspan $\gets$ 1 $+$ end\_row $-$ start\_row
		\State colspan $\gets$ 1 $+$ end\_col $-$ start\_col
		\For{row = start\_row \textbf{to} end\_row}
		\For{col $\gets$ start\_col \textbf{to} end\_col}
		\If{row $==$ start\_row \textbf{and} col $==$ start\_col}
		\State table[row][col] $\gets$ \{ "rowspan": rowspan,
		\State "colspan": colspan, "content": content \}
		\Else
		\State table[row][col] $\gets$ "merged"
		\EndIf
		\EndFor
		\EndFor
		\EndFor
	\end{algorithmic}
\end{algorithm}
\vspace{-0.08in}
In the second stage, we traverse the tabular data matrix row by row, gradually converting the stored logical information and cell content into a mark-up sequence, eventually generating the conversion result. The specific implementation is detailed in Algorithm \ref{algStage2}.

\vspace{-0.04in}
\begin{algorithm}
	\raggedright
	\small
	\caption{From Tabular Data Matrix to Markup Sequence}
	\textbf{Input:} table \\
	\textbf{Output:} markup
	\begin{algorithmic}[1]
		\State Initialize markup $\gets$ "\texttt{<table>}"
		\For{row \textbf{in} table}
		\label{algStage2}
		\State markup += "\texttt{<tr>}"
		\For{cell \textbf{in} row}
		\If{cell == "\texttt{merged}"}
		\State \textbf{continue}
		\Else
		\State rsp = cell["rowspan"]
		\State csp = cell["colspan"]
		\State markup += \texttt{<td rowspan} $= rsp \text{ colspan} = csp \texttt{>}$
		\State markup += cell["content"] + "\texttt{</td>}"
		\EndIf
		\EndFor
		\State markup += "\texttt{</tr>}"
		\EndFor
		\State markup += "\texttt{</table>}"
	\end{algorithmic}
\end{algorithm}

\section{Study of Table Size}

\begin{figure}[h] \centering
	\vspace{-0.04in}
	\includegraphics[width=0.23\textwidth]{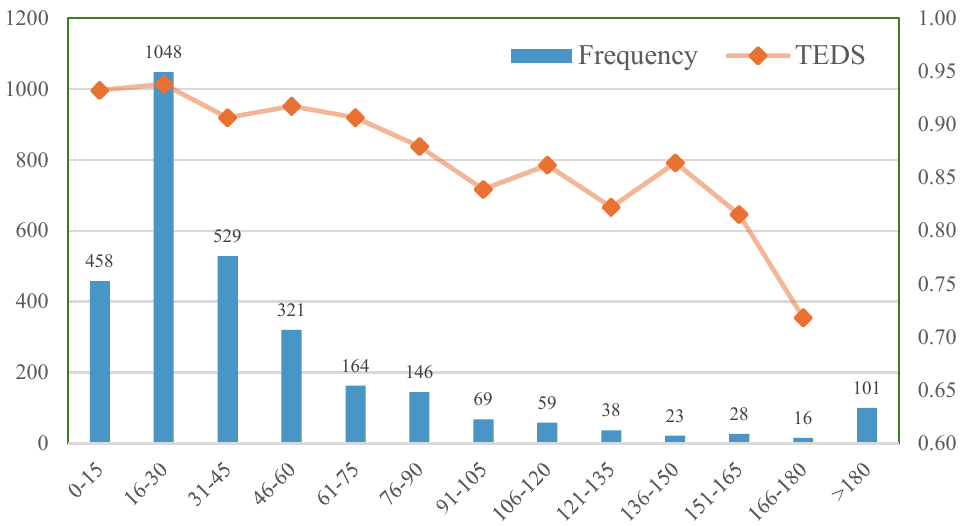}
	\includegraphics[width=0.23\textwidth]{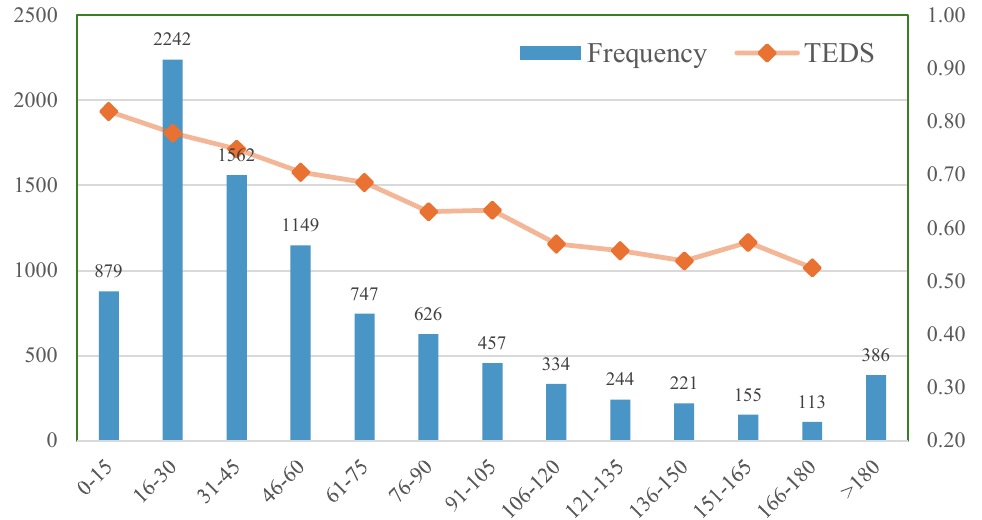}
	\vspace{-0.12in}
	\caption{Table size distribution and its impact on TR performance in SciTSR (left) and PubTabNet (right) Datasets}
	\vspace{-0.08in}
	\label{DatasetSzie}
\end{figure}

In this section, we analyze the scale of the tables involved in the benchmark evaluations. We quantify the size of tables in the SciTSR and PubTabNet datasets by counting the number of cells in each table. Furthermore, this section presents experiments designed to evaluate the impact of table size on table recognition performance. GPT is selected as a representative of VLLMs for these experiments, with the distribution details and experimental results illustrated in Figure \ref{DatasetSzie}. In small scale tables, the model demonstrates stable performance on both the SciTSR and PubTabNet datasets. However, as the table size increased, the model's performance experienced a moderate decline. This indicates that larger tables introduce longer context lengths, thereby affecting the VLLMs' table recognition task performance.

Specifically, statistical analysis shows that all tables involved in the experiments do not exceed the VLLMs' context window after being encoded into markup sequence. The largest table occupies only 10K tokens of context when encoded as an markup sequence, while all VLLMs possess context windows of at least 128K tokens. However, it is important to note that generating well formed markup sequence poses certain challenges to the model's capabilities. Therefore, when tables are huge and have complex structures, some VLLMs with weaker generative abilities may encounter issues generating parseable markup sequence results, even though the 128K context window is not exceeded.

\begin{figure}[h] \centering
	\vspace{-0.1in}
	\includegraphics[width=0.48\textwidth]{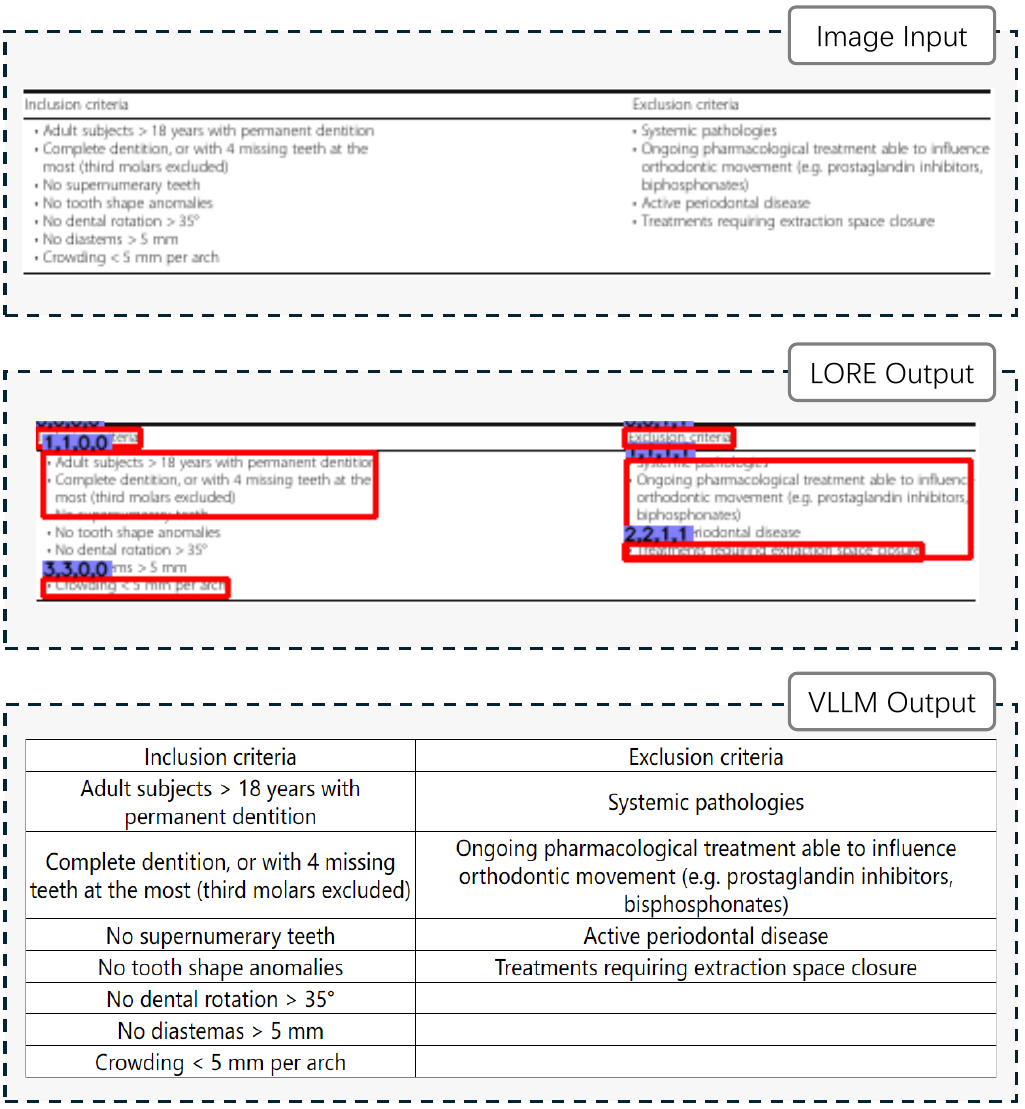}
	\vspace{-0.26in}
	\caption{A case study is conducted using a sample from PubTabNet.} \label{caseStudy1}
	\vspace{-0.12in}
\end{figure}

\vspace{-0.1in}
\section{Case Study}
Figure \ref{caseStudy1} presents a case study that showcases evaluation samples from the PubTabNet dataset and analyzes the table recognition results using the traditional OCR models LORE and VLLM. The input table image lacks clear row borders, making it challenging for traditional OCR models to locate specific cells accurately. However, VLLM effectively comprehends the hierarchical structure of the table, thereby producing correct recognition results. This case highlights the advantages of VLLM in semantically rich table recognition tasks, demonstrating its superior adaptability and robustness compared to traditional OCR models.

\begin{figure}[h] \centering
	\vspace{-0.1in}
	\includegraphics[width=0.48\textwidth]{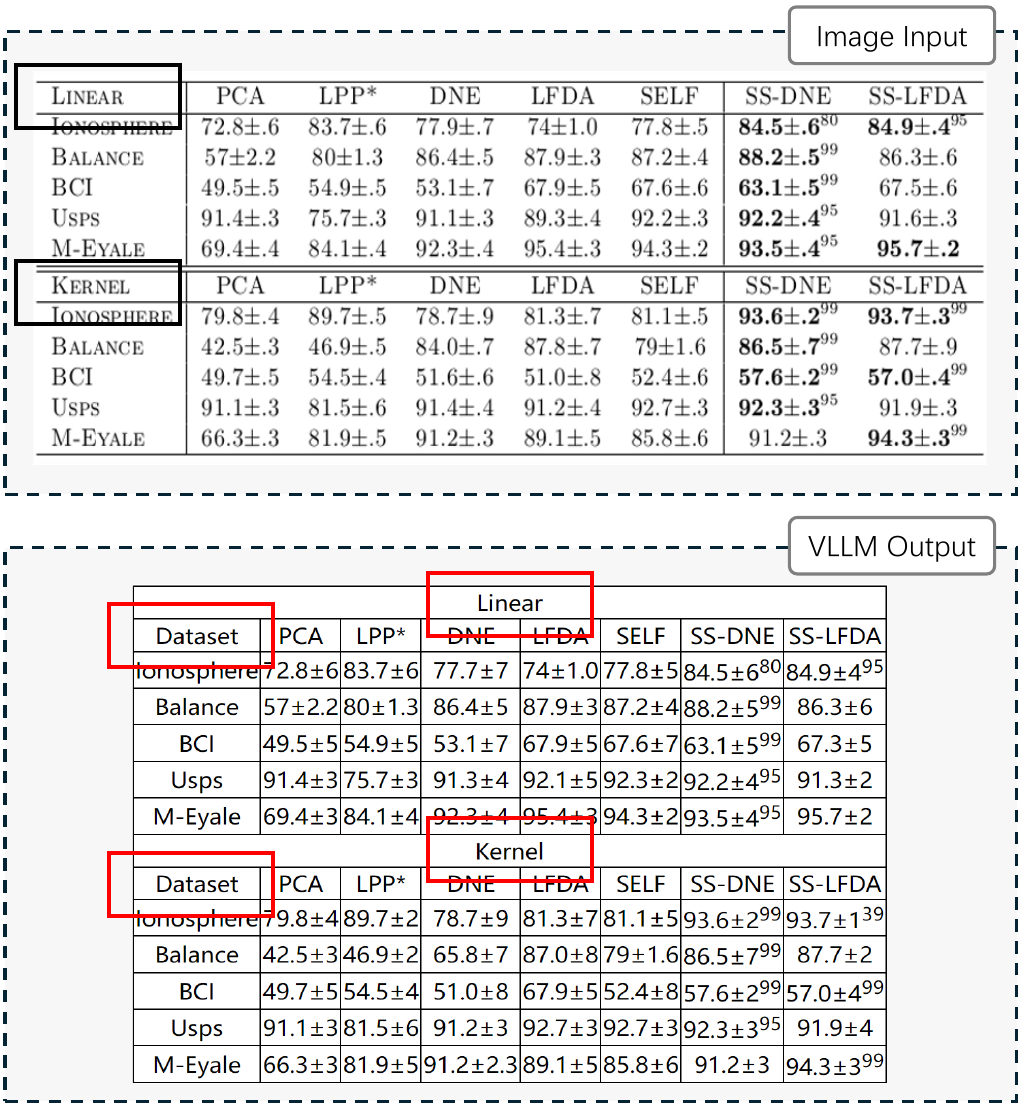}
	\vspace{-0.26in}
	\caption{A case study is conducted using a sample from SciTSR.} \label{caseStudy2}
	\vspace{-0.18in}
\end{figure}

However, relying on the understanding of the table's hierarchical structure to perform table recognition tasks can also have negative consequences in certain scenarios. As illustrated in Figure \ref{caseStudy2}, we analyzed the results of using VLLM for table recognition on evaluation samples from the SciTSR dataset. The results indicate that the output of VLLM in table recognition significantly deviates from the correct answers, particularly in the placement of the "Linear" and "Kernel" cells in the table. Specifically, VLLM tends to misidentify "Linear" and "Kernel" as belonging to the first row of the table, treating them as subheaders, even though these cells are located in the table's upper-left corner. 
This phenomenon may stem from VLLMs' excessive reliance on contextual cues, causing the model to be misled by its own reasoning or interpretation mechanisms when encountering atypical table structures, thereby resulting in erroneous hierarchical structural information recognition results.


\vspace{-0.1in}
\section{Parameter Settings of VLLMs} \label{vllm describe}

We report the parameter settings of six VLLMs. For open-source VLLMs, we select Phi-3.5-Vision-Instruct\footnote{\url{https://azure.microsoft.com/en-us/products/phi/}} (Phi) and Llama-3.2-90B-Vision-Instruct\footnote{\url{https://www.llama.com/}} (Llama) for evaluation. For closed-source VLLMs, we evaluate GPT-4o-mini (GPT-mini), Qwen-VL-Max\footnote{\url{https://qwenlm.github.io/blog/qwen-vl/}} (Qwen), GPT-4o\footnote{\url{https://openai.com/index/hello-gpt-4o/}} (GPT) and Gemini-1.5-Pro \footnote{\url{https://deepmind.google/technologies/gemini/pro/}} (Gemini). Specifically, when generating multiple tool invocation plans, we set the temperature parameter to 0.8 to encourage the generation of more diverse tool invocation plans. For other experiments, the temperature parameter was kept at 0 to ensure the stability of the experimental results. Additionally, the top\_p parameter was set to 0.2, and the n\_samples parameter was set to 1.

\vspace{-0.1in}
\section{Details of TEDS Metric} \label{teds calculate}
TEDS is designed to measure the similarity between two tree structures. Specifically, TEDS calculates the similarity between the HTML tree of the real tag and the HTML tree of the predicted tag. When applying the TEDS metric, the HTML table format must first be converted into a tree structure. The similarity is then calculated using the following formula:
\begin{equation}
        \setlength{\abovedisplayskip}{3pt}
	\setlength{\belowdisplayskip}{3pt}
	\text{TEDS}(T_a, T_b) = 1 - \frac{\text{EditDist}(T_a, T_b)}{\max(|T_a|, |T_b|)},
\end{equation}
where \( T_a \) and \( T_b \) represent the tree structures of the tables in HTML format. \(\text{EditDist}(T_a, T_b)\) denotes the tree-edit distance, and \( |T| \) refers to the number of nodes in tree \( T \).
Additionally, we adopted a modified version of TEDS, referred to as TEDS-Struct. TEDS-Struct is designed to assess the accuracy of table structure recognition without considering the specific results generated by the table content.

\vspace{-0.1in}
\section{Datasets}

We selected three public datasets to evaluate our model's performance across various table structures and conditions. These datasets provide a well-rounded basis for testing and comparing table recognition methods.

\noindent \textbf{SciTSR} is a dataset of the scientific literature containing 15,000 tables and their structural labels from LaTeX source files. It also has 12,000 training samples and 3,000 test samples. The images are high-quality and include complex tables.

\noindent \textbf{PubTabNet} is created by matching XML files with PDF files of scientific articles from the Open Access Subset of PubMed Central™. PubTabNet includes 500,777 training samples and 9,115 validation samples with their HTML representations. PubTabNet is for end-to-end model training, where the table area is turned into an image with a resolution of 72 pixels per inch (PPI). This low PPI reduces the computational load but lowers image quality.

\noindent \textbf{WTW} is a dataset of real-world scenes with 10,970 training samples and 3,611 test samples. It includes structural information for each table and extreme cases of real-world tables, such as tilt, overlap, blur, occlusion, and curvature. WTW focuses on recognizing structural information and does not provide labels for cell content.

\begin{table}[t]\centering
        \vspace{0.1in}
	\caption{Characteristics of selected table recognition datasets.}
	\vspace{-0.1in}
	\label{tab:dataset}
	\resizebox{0.48\textwidth}{!}{
		\large
		\renewcommand{\arraystretch}{1.25}
		\begin{tabular}{*{10}{c}}
			\toprule
			Dataset  & Tables & Source & \begin{tabular}[c]{@{}c@{}}Markup Sequence\\ Annotations\end{tabular} & \begin{tabular}[c]{@{}c@{}}Textual\\ Annotations\end{tabular} & Year \\
			\midrule
			SciTSR       & 15K    & ArXiv              & $\times$ & $\checkmark$ & 2019 \\
			PubTabNet    & 568K   & PubMed             & $\checkmark$ & $\checkmark$ & 2020 \\
			WTW          & 14.6K  & Wild Scenarios     & $\times$ & $\times$ & 2021 \\
			\bottomrule
		\end{tabular}
	}
	\vspace{-0.15in}
\end{table}

\section{Implementation Details of Benchmark}
\begin{figure}[h] \centering
        \vspace{-0.12in}
	\includegraphics[width=0.48\textwidth]{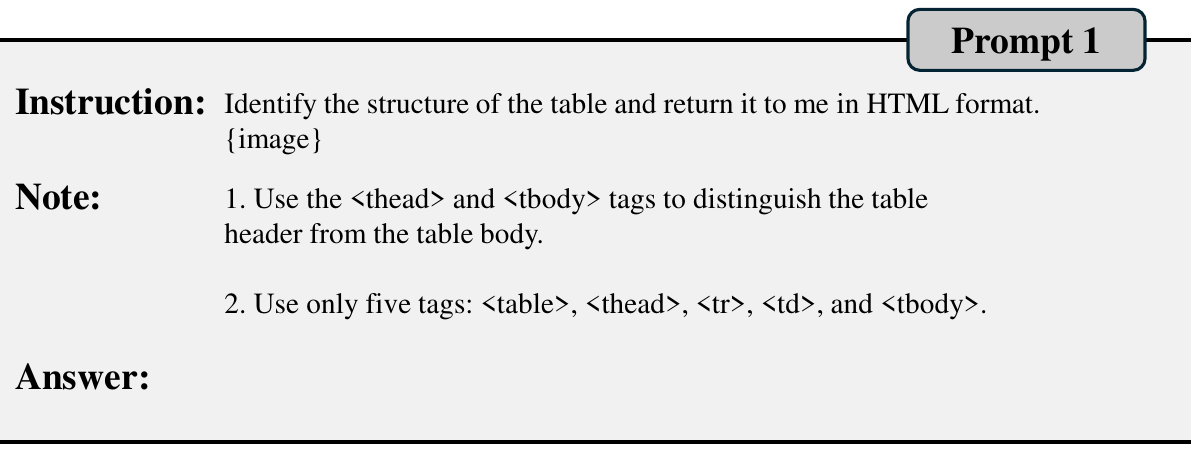}
	\vspace{-0.12in}
	\captionsetup{labelformat=default, labelsep=colon, name=Prompt}
	\vspace{-0.12in}
	\caption{Simplified prompt template for VLLMs on table recognition tasks.}
	\vspace{-0.12in}
	\label{prompt:1}
\end{figure}

\begin{figure}[h] \centering
	\vspace{-0.15in}
	\includegraphics[width=0.48\textwidth]{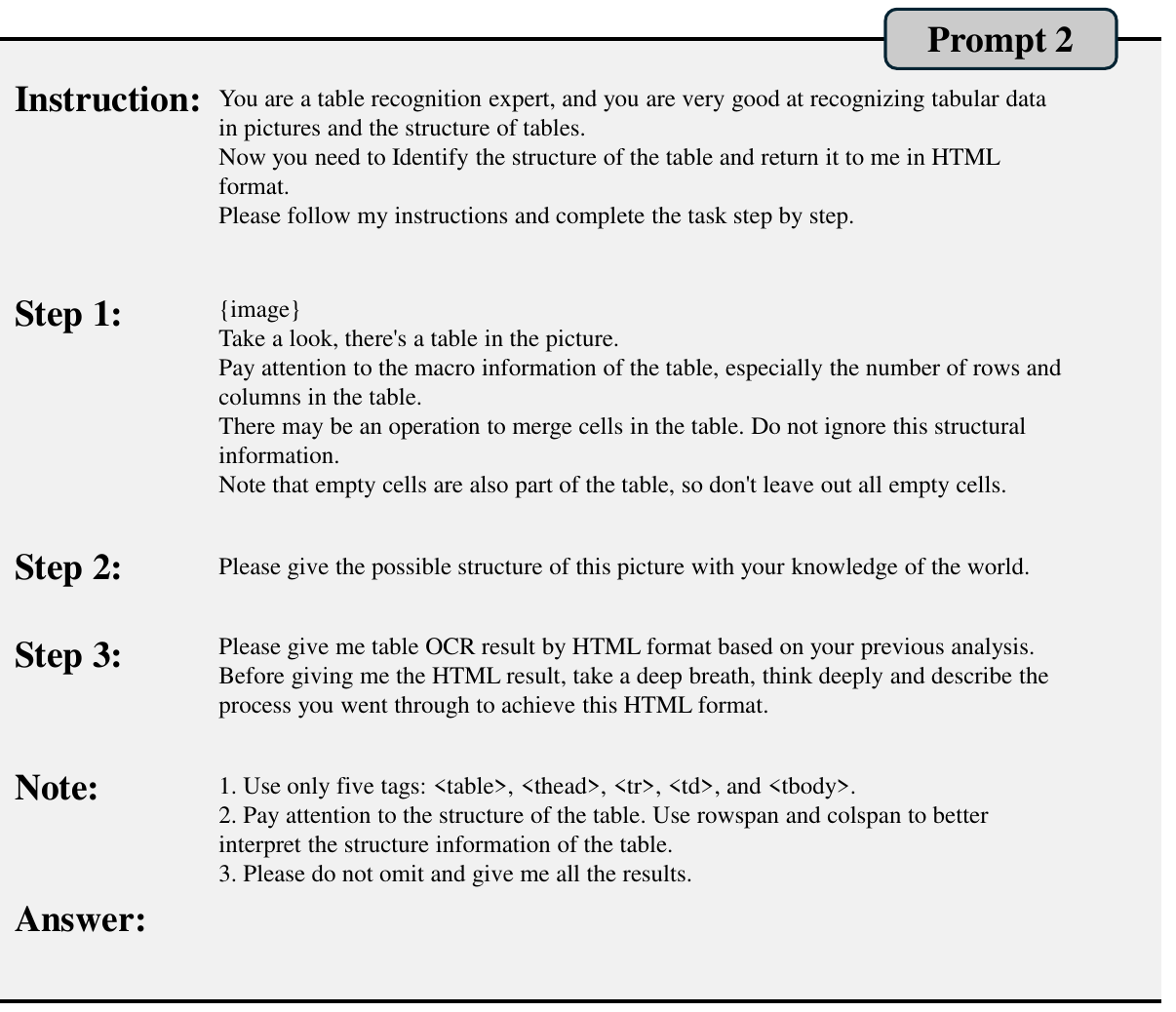}
	\caption{Chain-of-Thought prompt template for VLLMs on table recognition tasks.}
	\label{prompt:2}
        \vspace{-0.1in}
\end{figure}

We present two prompt templates designed to guide VLLM in generating markup sequence outputs for TR tasks in Figures~\ref{prompt:1} and~\ref{prompt:2}. Prompt 1 is a simplified template that facilitates the execution of TR tasks by VLLM. In contrast, Prompt 2 introduces the concept of Chain-of-Thought by explicitly planning the table recognition process for VLLM. In our experiments, the more complex Prompt 2 achieved a slight improvement in the performance of table recognition tasks. However, since our benchmark evaluation aims to reflect VLLM's recognition capabilities more directly, we employ Prompt 1 as the prompt template in all experiments.

\begin{figure}[tp] \centering
        \vspace{0.1in}
	\includegraphics[width=0.48\textwidth]{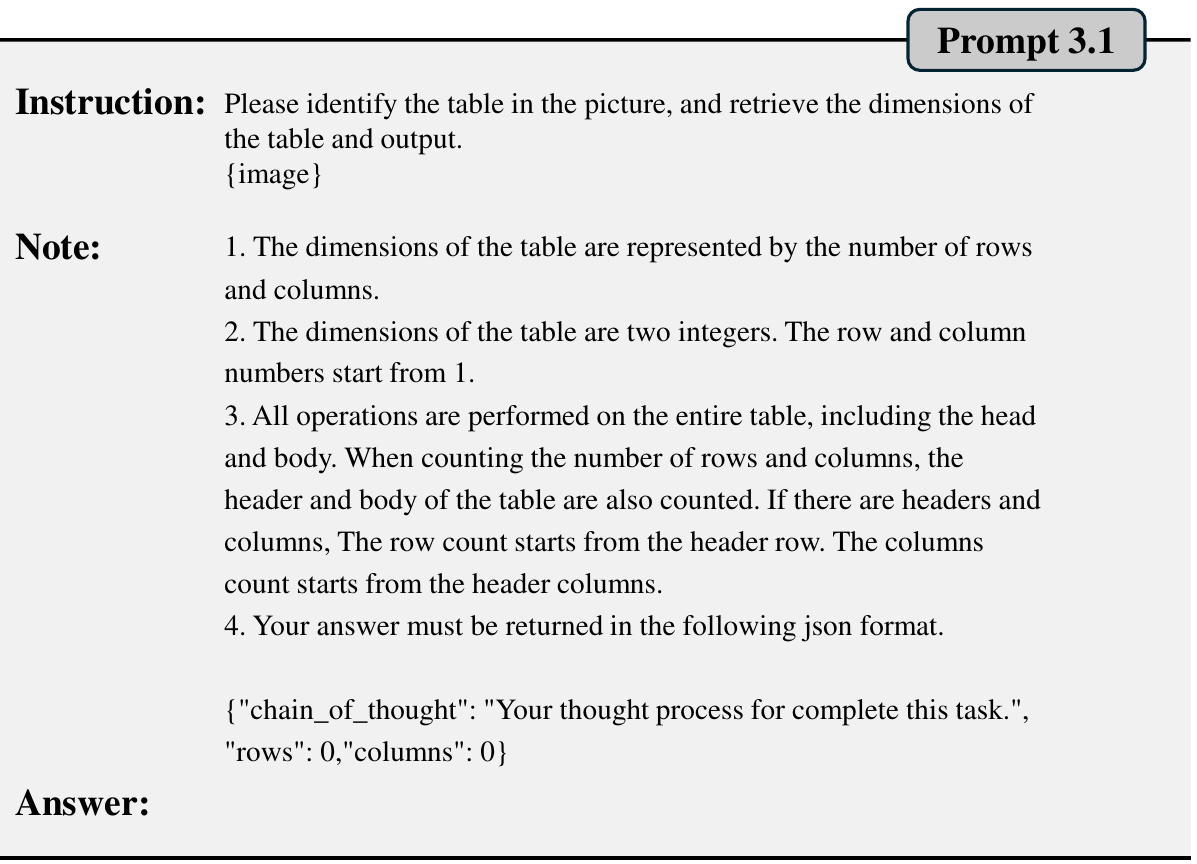}
	\caption{Prompt template on visual table size detection task.}
	\label{prompt:31}
	\vspace{-0.02in}
\end{figure}

\begin{figure}[tp] \centering
	\vspace{-0.1in}
	\includegraphics[width=0.48\textwidth]{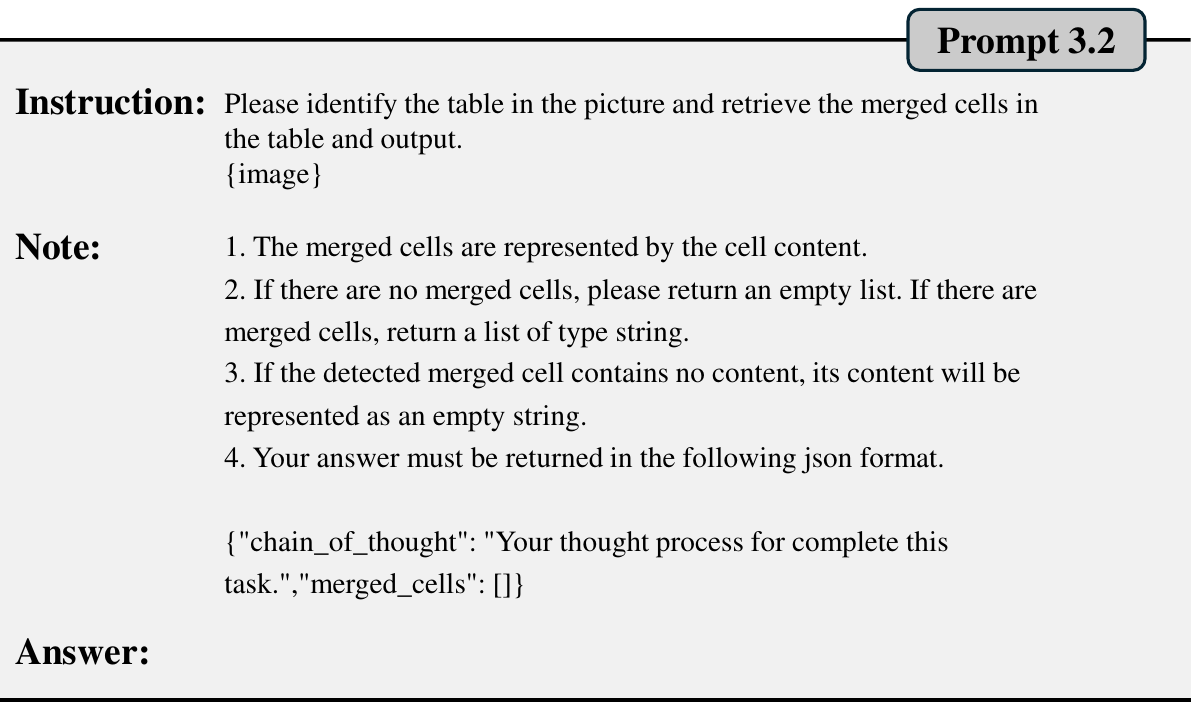}
	\vspace{-0.22in}
	\caption{Prompt template on merged cell detection task.}
	\vspace{-0.1in}
	\label{prompt:32}
\end{figure}

\begin{figure}[htp] \centering
        \vspace{-0.04in}
	\includegraphics[width=0.48\textwidth]{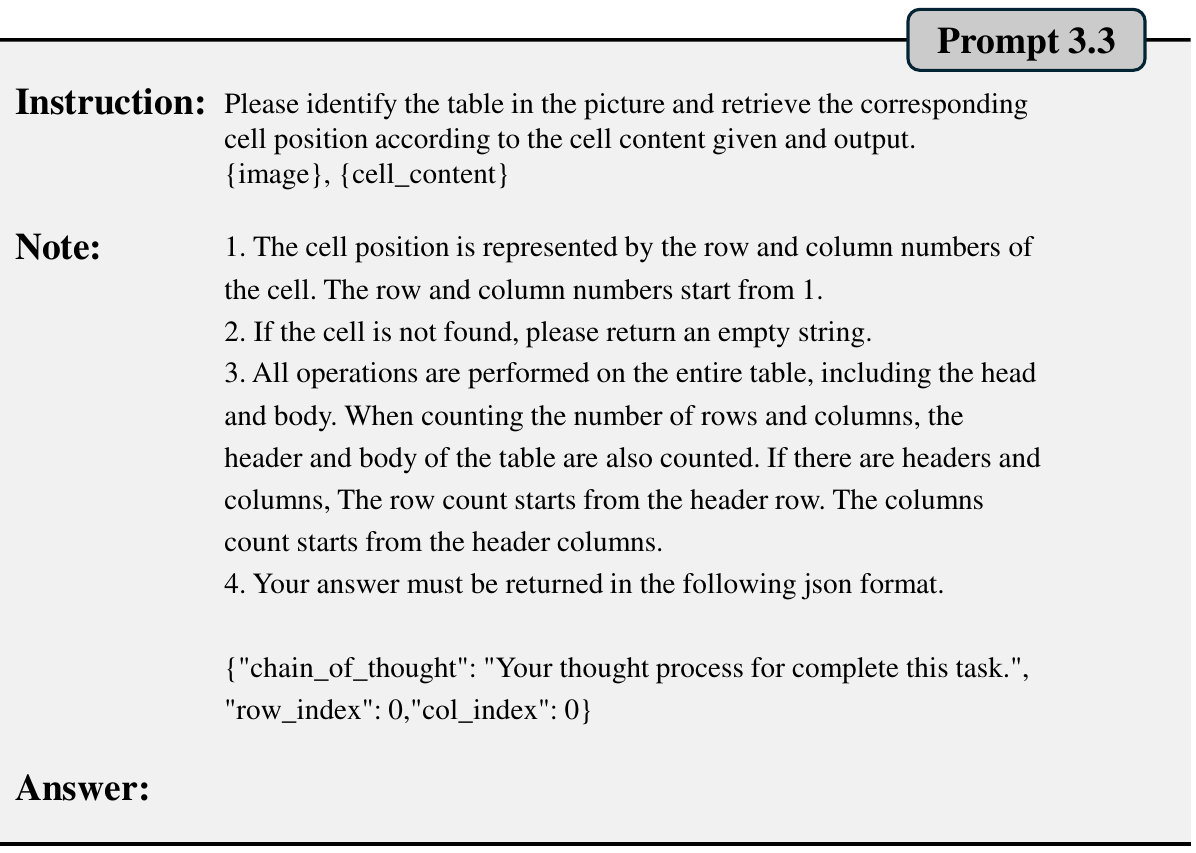}
	\vspace{-0.12in}
	\caption{Prompt template on content-based cell recognition task.}
	\vspace{-0.12in}
	\label{prompt:33}
\end{figure}

\begin{figure}[htp] \centering
	\vspace{-0.04in}
	\includegraphics[width=0.48\textwidth]{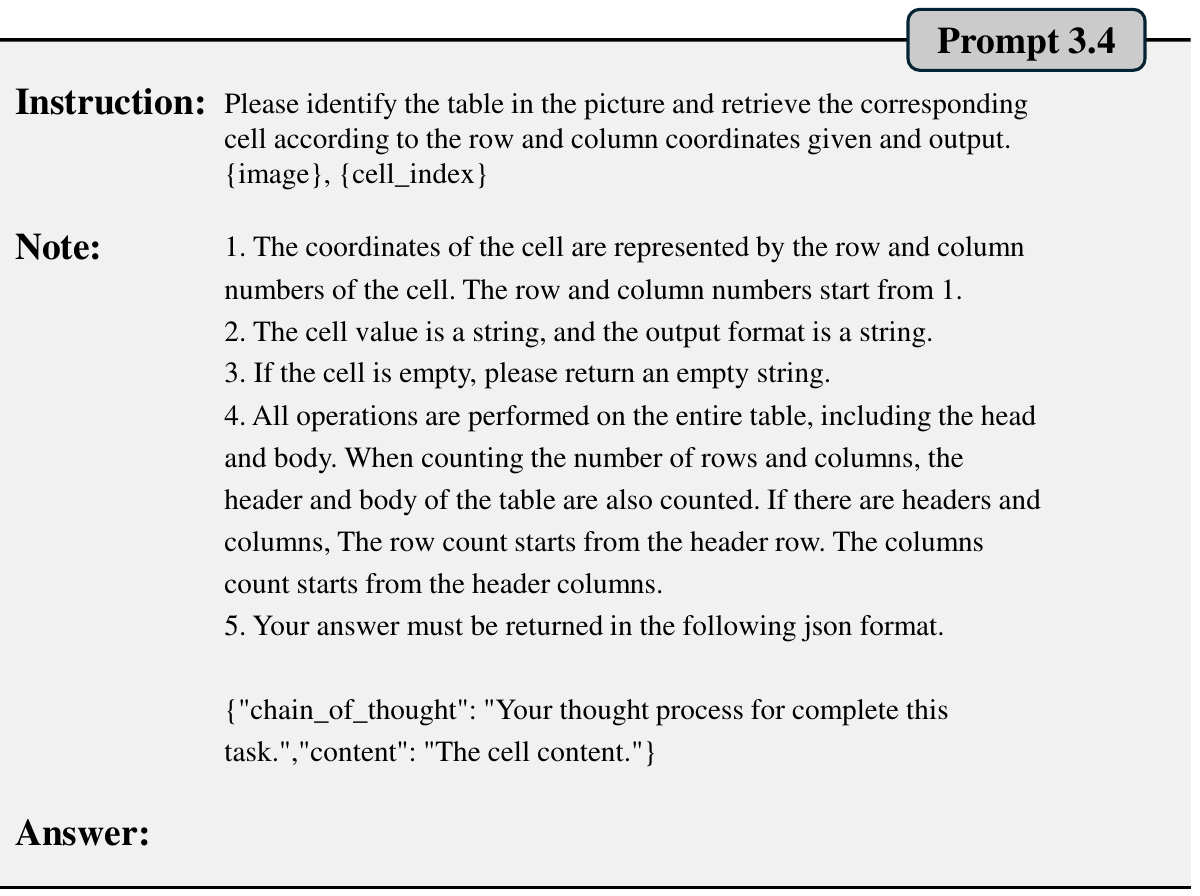}
	\vspace{-0.12in}
	\caption{Prompt template on index-based cell recognition task.}
	\vspace{-0.2in}
	\label{prompt:34}
\end{figure}

\begin{figure}[htp] \centering
	\includegraphics[width=0.48\textwidth]{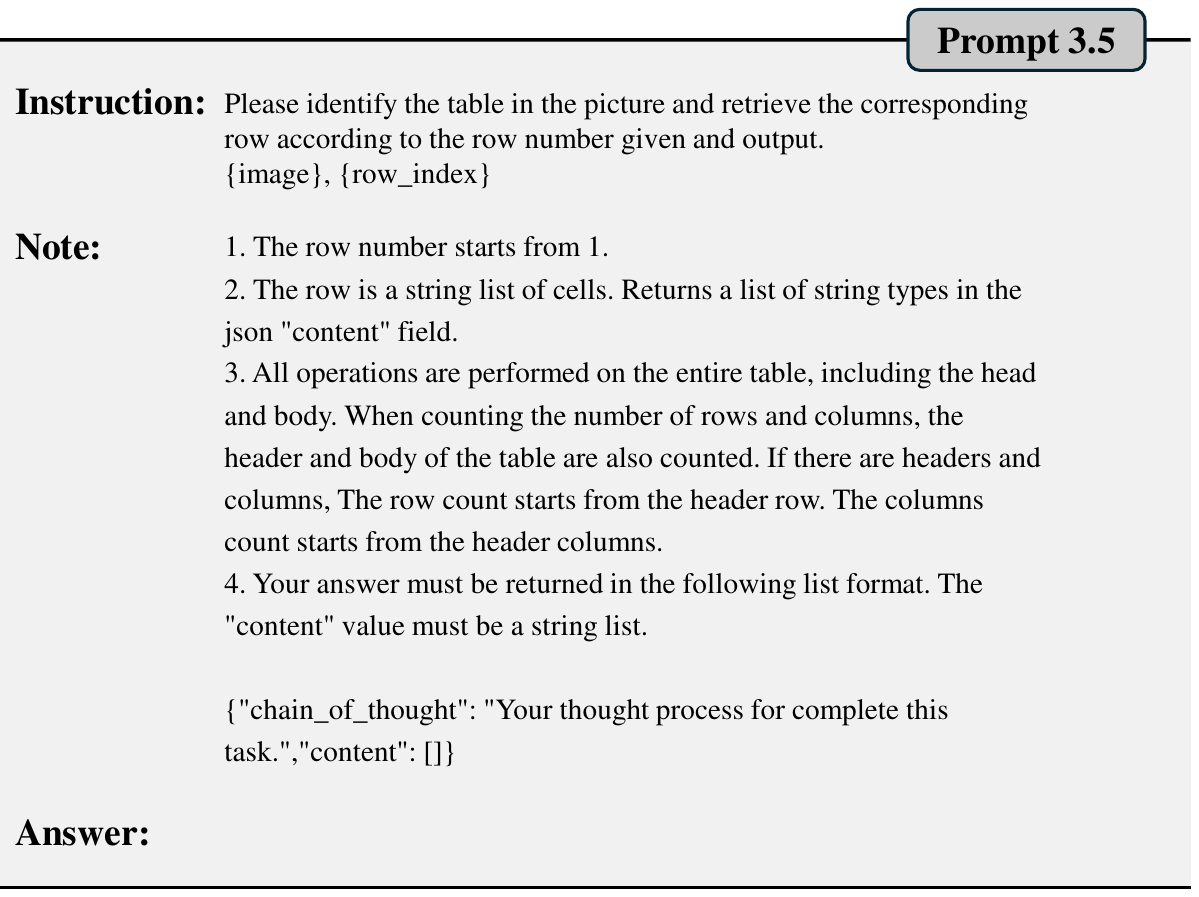}
	\vspace{-0.1in}
	\caption{Prompt template for row index recognition task.}
	\vspace{-0.12in}
	\label{prompt:35}
\end{figure}

\begin{figure}[htp] \centering
        \vspace{0.1in}
	\includegraphics[width=0.48\textwidth]{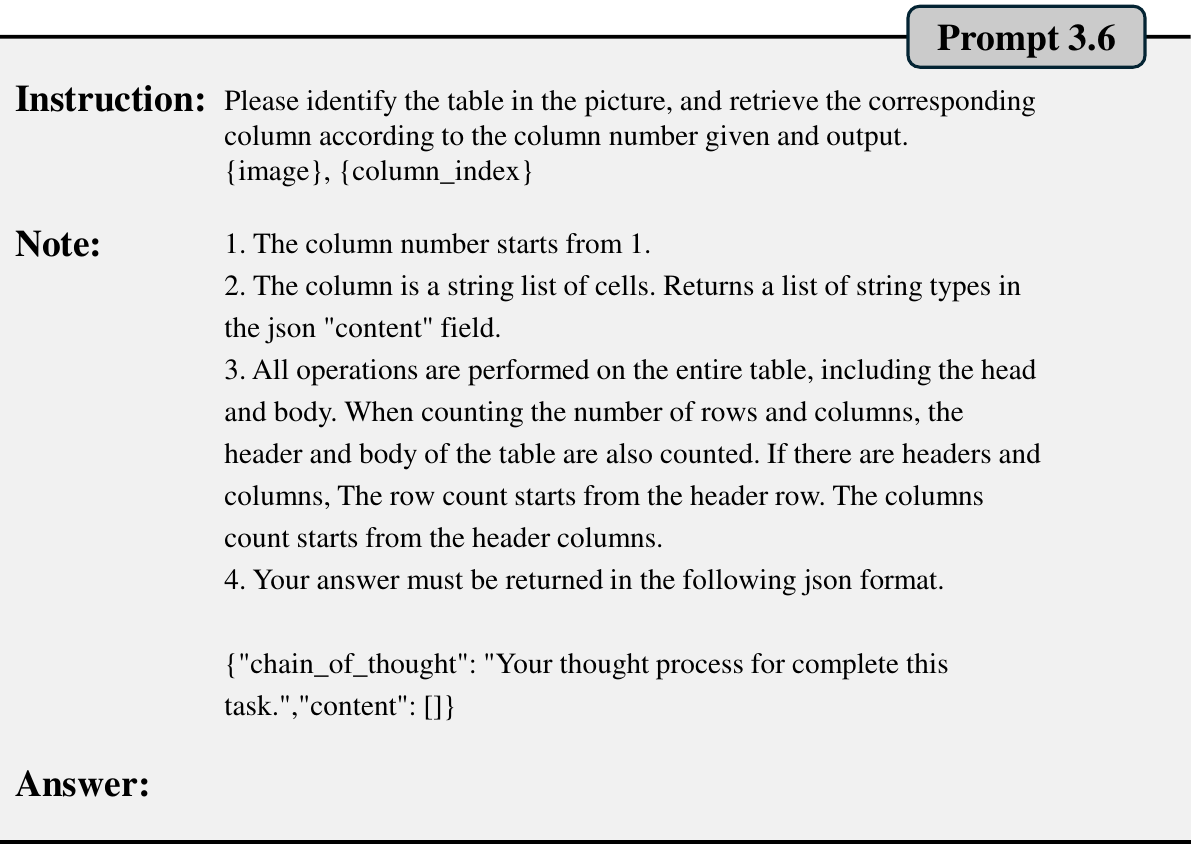}
	\vspace{-0.1in}
	\caption{Prompt template for column index recognition task.}
	\vspace{-0.06in}
	\label{prompt:36}
\end{figure}

Figures \ref{prompt:31}-\ref{prompt:36} illustrate the prompt templates designed for our benchmark evaluation of hierarchical evaluation tasks. Each template corresponds to a specific hierarchical task:

\begin{itemize}\setlength{\itemsep}{3pt}
	\item Visual Table Size Detection Task. (Figure \ref{prompt:31})
	\item Merged Cell Detection Task. (Figure \ref{prompt:32})
	\item Content-based Cell Recognition Task. (Figure \ref{prompt:33})
	\item Index-based Cell Recognition Task. (Figure \ref{prompt:34})
	\item Index-based Row Recognition Task. (Figure \ref{prompt:35})
        \item Index-based Line Recognition Task. (Figure \ref{prompt:36})
\end{itemize}

\begin{figure}[htp] \centering
	\vspace{-0.06in}
	\includegraphics[width=0.48\textwidth]{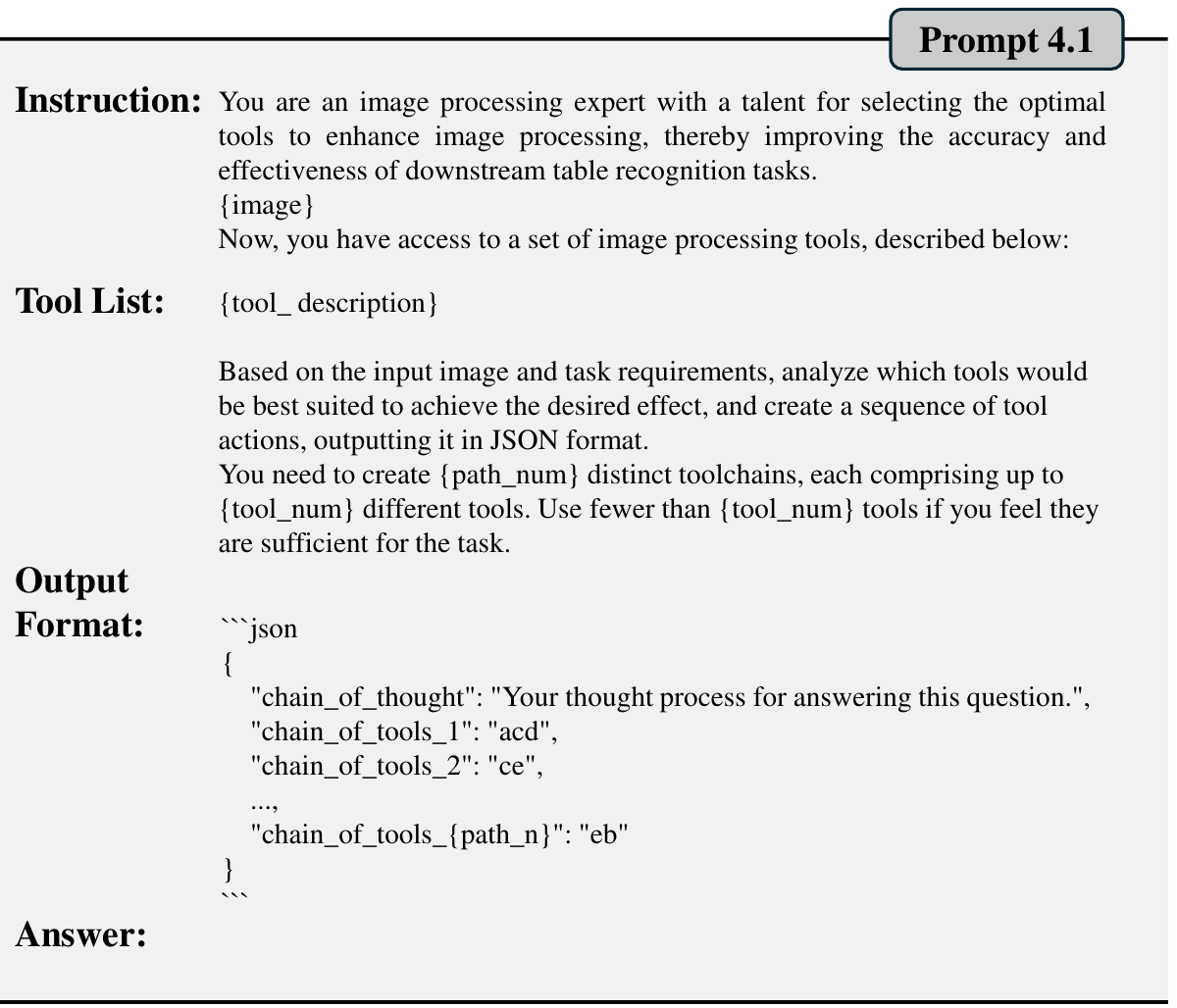}
	\vspace{-0.1in}
	\caption{Prompt template for generating multiple tool invocation plans in the NGTR framework}
	\vspace{-0.08in}
	\label{prompt:41}
\end{figure}

\begin{figure}[htp] \centering
        \vspace{0.1in}
	\includegraphics[width=0.48\textwidth]{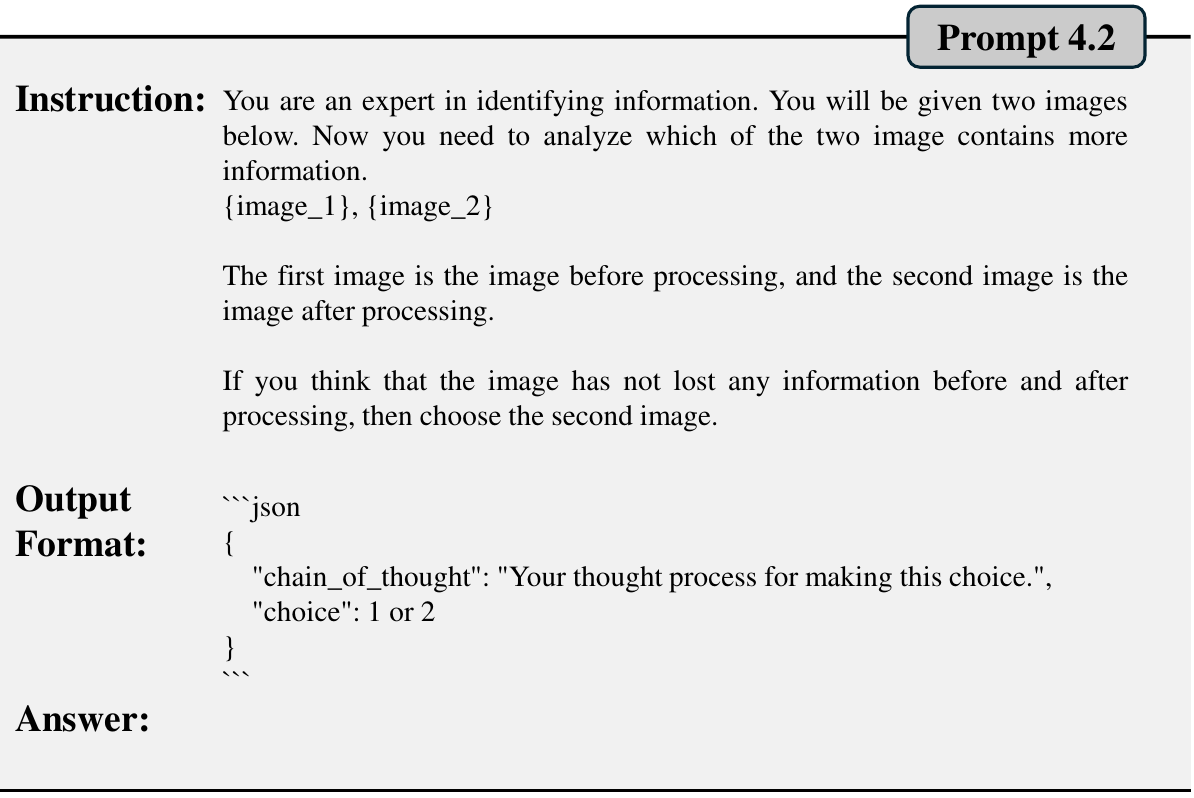}
	\vspace{-0.1in}
	\caption{Prompt template for the reflection-driven tool utilization module in the NGTR framework}
	\vspace{-0.08in}
	\label{prompt:42}
\end{figure}

\vspace{-0.1in}
\section{Implementation Details of NGTR}
Figure \ref{prompt:41} illustrates the prompt template utilized by our proposed NGTR framework for generating multiple tool invocation plans. By inputting the target image along with its tool descriptions and identifiers, we leverage the planning capabilities of VLLM to produce several tool invocation plans for subsequent modules. This module embeds two adjustable hyperparameters into the prompt template for process adjustment. The designed hyperparameters include the maximum length \( L \) of the toolchain execution plan and the number \( N \) of plans generated per iteration.

Figure \ref{prompt:42} illustrates the prompt template employed by the NGTR framework within the reflection-driven tool utilization module. By inputting both the pre-processed and post-processed images, we could leverage the discriminative capabilities of VLLM to select the image with superior quality. This selection process enables the further optimization of tool invocation strategies, thereby minimizing the loss of image information caused by inappropriate tool usage.


\end{document}